\documentclass[letterpaper]{article}

\usepackage{helvet} 
\usepackage{courier} 

\usepackage{amsmath}
\usepackage{graphicx}
\usepackage{textcomp}
\usepackage{xcolor}
\usepackage{multirow}

\usepackage{subcaption}

\usepackage[linesnumbered,ruled,vlined]{algorithm2e}
\SetKwRepeat{Do}{do}{while}%

\newcommand{\bluecomment}[1]{\textcolor{blue}{#1}}

\newcommand{\Input}[1]{\textbf{Input:} #1\\}
\newcommand{\Output}[1]{\textbf{Output:} #1\\}

\title{P2W: From \underline{P}ower Traces \underline{to} \underline{W}eights Matrix - \\ An Unconventional Transfer Learning Approach}

\author{
    \textbf{Roozbeh Siyadatzadeh}\thanks{s.r.siyadatzadeh@liacs.leidenuniv.nl}, \textbf{Fatemeh Mehrafrooz},\\
    \textbf{Nele Mentens}, and \textbf{Todor Stefanov} \\
    Leiden Institute of Advanced Computer Science (LIACS) \\
    Leiden University, The Netherlands
}

\date{}

\begin{document}
\maketitle

\begin{abstract}
The rapid growth of deploying machine learning (ML) models within embedded systems on a chip (SoCs) has led to transformative shifts in fields like healthcare and autonomous vehicles. One of the primary challenges for training such embedded ML models is the lack of publicly available high-quality training data. Transfer learning approaches address this challenge by utilizing the knowledge encapsulated in an existing ML model as a starting point for training a new ML model. However, existing transfer learning approaches require direct access to the existing model which is not always feasible, especially for ML models deployed on embedded SoCs. Therefore, in this paper, we introduce a novel unconventional transfer learning approach to train a new ML model by extracting and using weights from an existing ML model running on an embedded SoC without having access to the model within the SoC. Our approach captures power consumption measurements from the SoC while it is executing the ML model and translates them to an approximated weights matrix used to initialize the new ML model. This improves the learning efficiency and predictive performance of the new model, especially in scenarios with limited data available to train the model. Our novel approach can effectively increase the accuracy of the new ML model up to 3 times compared to classical training methods using the same amount of limited training data.
\end{abstract}


\section{Introduction}
The advent and subsequent evolution of machine learning (ML) technologies have impacted various aspects of modern life, notably in fields like healthcare, autonomous vehicles, and cybersecurity \cite{Sarker2023}. However, one of the primary challenges in the ML domain is the gathering of relevant, high-quality data for training an ML model for a specific task. This is because of several reasons such as data confidentiality, high cost, time, and effort of data gathering \cite{10.1007/978-3-030-01364-6_17}. One way to address this challenge is to use the knowledge encapsulated in existing ML models that are used for another related task, and to optimize and fine-tune these models to make them suitable for our new task. Such process is known as transfer learning~\cite{10172347}.

Despite the clear benefits of transfer learning, a significant issue persists: it is not always easy to find a publicly available well-trained model to start from. For example, it is relatively easy to find a model on the internet for classifying simple objects, but it is not as easy to find a well-trained model for some classes of medical tasks due to several reasons, such as privacy and the expenses associated with such models. This lack of publicly available well-trained models for a specific class of tasks is even more severe in the domain of embedded systems on a chip (SoCs) for ML. For such SoCs, the design of a suitable ML model is impacted by limitations in resources and other requirements to make the model suitable for a specific kind of embedded SoC. These factors often prevent the full realization of the potential benefits of transfer learning in the embedded SoC domain.

Although public availability of well-trained ML models suitable for transfer learning in the embedded systems domain is limited, there are embedded SoCs everywhere that are running such ML models for a wide variety of tasks. The problem is that the general public does not have direct access to the models inside these chips and cannot benefit from the model's knowledge by utilizing conventional transfer learning approaches that require direct access to ML models. 

Therefore, in this paper, we present a proof-of-concept of a novel unconventional transfer learning approach called \textbf{P2W}. Our approach facilitates the transfer of knowledge encapsulated in a relevant well-trained neural network, running inside an embedded SoC, to a new neural network. P2W enables effective training of the new neural network model to perform new similar tasks. This is done by measuring the power consumption of the SoC while it is running the ML model. We take multiple power traces with different input data and analyze them with the help of an Encoder-Decoder Deep Neural Network (EDNN).
Through such power trace analysis we extract, to a large extent, the knowledge of the well-trained ML model in the form of an \emph{approximated} weights matrix. This matrix serves as foundational knowledge to start training new ML models. By initializing the new models with the approximated weights matrix (i.e., transferring the knowledge) and further training them with a limited amount of data, we obtain new well-trained ML models that can perform the same or related tasks very well.

The main novel contributions of the paper can be summarized as follows:

\begin{itemize}
\item We propose an unconventional transfer learning approach which circumvents the traditional barriers for researchers/developers of ML models related to the acquisition of high-quality data for training, in sensitive domains such as healthcare for example, where data availability is often restricted due to privacy/ethical concerns. By extracting encoded knowledge in the form of an approximated weights matrix, purely from power consumption traces, researchers/developers can transfer this knowledge to new models without access to sensitive information.

\item To support our approach, we introduce a new power analysis technique that utilizes an \texttt{EDNN} to translate power traces into an approximated weights matrix.

\item Given a limited training dataset, we evaluate the effectiveness of our P2W transfer learning approach by comparing the accuracy of an ML model trained with and without the use of P2W. Our experimental results show that after training the ML model with our P2W approach, we are able to improve the initial average accuracy of the model, which is 37\% using only the limited training dataset, to 97\% with the aid of P2W followed by fine-tuning with the same dataset.
\end{itemize}

The remainder of this paper is organized as follows.
Section~\ref{sec:RelatedWork} discusses related work, after which Section~\ref{sec:Background} provides an overview of some existing concepts on which our P2W approach relies. Section~\ref{sec:P2W} 
presents our P2W transfer learning approach in details. In Section~\ref{sec:Evaluation}, we evaluate our approach based on several metrics to demonstrate its practical applicability and usefulness. Finally, Section~\ref{sec:Conclusions} concludes this paper.

\section{Related Work}
\label{sec:RelatedWork}
This section provides an overview of the most relevant related work, divided into three distinct categories: Power Analysis techniques, Machine Learning on Embedded Systems, and Transfer Learning approaches.

\emph{Power Analysis:}
Power analysis techniques have been pivotal in reverse engineering of neural networks, where researchers employ timing, power, and electromagnetic (EM) data to reveal intricate details like activation functions and network structures \cite{10.5555/3361338.3361374}. Wang et al. \cite{10077551} explore how to deduce the structure of deep neural network (DNN) models deployed within in-memory computing (IMC) systems through power analysis. Investigations in \cite{9000972} collect voltage and current data to infer model structures, and \cite{8735505} demonstrate an attack on FPGA-based DNN accelerators using EM leakage. In~\cite{10.1145/3195970.3196105}, the authors present a novel approach of utilizing cache timing side channels that offers insights into DNN structures. Our new power analysis technique complements the aforementioned techniques by generating an approximated weights matrix directly from power traces, which the existing techniques do not provide, as they focus mainly on uncovering the DNN model structure. In contrast, our focus is on transferring as much as possible knowledge, encapsulated in a DNN, via the approximated weights matrix. 

\emph{Machine Learning on Embedded Systems:}
ML models deployed on embedded systems find use in a wide range of fields, including medical imaging and autonomous driving. Innovative applications of ML models include GFUNet, which integrates Fourier Transform with U-Net architecture for medical image segmentation \cite{LI2023107290}. Edge computing benefits from a long short-term memory (LSTM) model on TensorFlow Lite for gesture recognition in wearable devices as demonstrates in~\cite{9399005}. A lightweight combination of LSTM and a multi-layer perceptron (MLP) model has enabled real-time electrocardiogram (ECG) anomaly detection for internet of things (IoT) devices \cite{9669005}. Additionally, MLP models have been optimized for medical decision-making within medical IoT (MIoT) using boosting and dimension reduction for efficient predictions \cite{LEE202015}. Tiny convolutional neural networks (CNNs) on low-power devices have demonstrated the robustness of transfer learning in autonomous driving~\cite{Rana_2023}, while MLP models have shown efficacy in early breast cancer detection~\cite{Rana_2023}. In contrast, our work harnesses the power of \texttt{EDNN} models to perform analysis of power traces captured for SoCs, thereby enabling the generation of approximated weights matrices for transfer learning, which is a novel application of \texttt{EDNNs} in the transfer learning and embedded systems domain.

\emph{Transfer Learning:}
Transfer learning addresses the challenge of scarce or difficult-to-obtain training data for ML models by leveraging data from a related but different source domain, thereby improving models' performance even when traditional assumptions of identical feature spaces and data distributions do not hold \cite{Weiss2016}. For instance, in text sentiment classification, transfer learning can enhance predictions when training data from one domain (e.g., digital camera reviews) are used for another domain (e.g., food reviews) \cite{5288526}. In image classification, models pre-trained on large datasets can be fine-tuned for specific tasks with limited data, such as adapting general image recognition models to medical image analysis \cite{10.1109/CVPR.2011.5995702, Zhu_Chen_Lu_Pan_Xue_Yu_Yang_2011}. Human activity recognition benefits from transferring knowledge across different sets of activities, improving model accuracy when applied to new activities \cite{10.5555/3104482.3104533}. In software defect detection, transfer learning allows models trained on data from one software project to be applied to another, even with different metrics \cite{10.1145/2786805.2786814}. Additionally, multi-language text classification uses transfer learning to adapt models for different languages, enabling improved performance in scenarios with limited labeled data in the target language \cite{JMLR:v20:13-580, Zhou_Pan_Tsang_Yan_2014}. Other applications include environmental monitoring, where models trained on data from one geographical region are adapted to another, and financial forecasting, where models utilize data from different market conditions to enhance predictive accuracy.

These transfer learning approaches typically require direct access to a fully functional ML model and its parameters and coefficients, which is not always feasible. In contrast to existing transfer learning approaches, our P2W transfer learning approach does not require direct access to such parameters and coefficients because it utilizes power traces to extract such information. Thus, P2W expands the applicability of transfer learning to scenarios where direct access to ML models is not feasible.

\section{Background}
\label{sec:Background}
In our P2W transfer learning approach, we utilize a variety of methods and models, and in this section, we provide an overview of the three most crucial among them. These include the \emph{Power Analysis} method, the \emph{EDNN} model, and \emph{F1-score}.

\subsection{Power Analysis}
\label{sec:PowerAnalBackground}
Power analysis is a method that exploits variations in the power consumption of a processing system during operation. The method analyzes the power consumption pattern to extract (sensitive) internal data and computational states~\cite{10.1007/3-540-48405-1_25}. Initially developed for probing cryptographic systems, power analysis has now expanded its applications, finding relevance in fields such as neural networks~\cite{Real2021PhysicalSA}.

Differences in the power consumption patterns are at the core of power analysis. For instance, it is possible to monitor the power consumption while a device performs cryptographic operations and use this data to infer secret keys or internal states. The power consumption $ P $ at any time $ t $ can be expressed as:
\begin{equation}
P(t) = \sum_{i=1}^{n} a_i \cdot P_i(t) + P_{\text{static}}
\end{equation}
where $ a_i $ represents the activity factor of the $ i $-th internal component, $ P_i(t) $ is the dynamic (data-dependent) power consumption, and $ P_{\text{static}} $ is the static power consumption. By observing the variance in $ P(t) $, particularly during data-dependent operations, it is possible to discern patterns that correlate with the processed data~\cite{10.1007/3-540-48405-1_25}.

\subsection{Encoder-Decoder Deep Neural Network (EDNN)}
\label{sec:AutoEncBackground}
An Encoder-Decoder Deep Neural Network (EDNN), is a specialized form of a deep neural network designed to learn efficient representations (or encodings) of input data. The architecture of an \texttt{EDNN} is divided into two main parts: the \emph{encoder}, which condenses the input data into a compressed representation or code, and the \emph{decoder}, which attempts to reconstruct the input data from this compressed code.

The process of training an \texttt{EDNN} is fundamentally centered around minimizing a loss function, which quantifies the discrepancy between the original input data and the generated data produced by the network. The loss function can be expressed as follows~\cite{nielsen2015neural}:

\begin{equation}
\label{Eq:loss_function}
\text{Loss}(X, \hat{X}) = \frac{1}{N} \sum_{i=1}^{N} \|f(x_i; \phi) - \hat{x}_i\|^2
\end{equation}

In Equation~\ref{Eq:loss_function}, $X$ and \(\hat{X}\) denote the sets of input and target data respectively, where each $x_i \in X$ is an individual input data point and $\hat{x}_i \in \hat{X}$ is its corresponding target. The function $f(x_i; \phi)$ represents the output of the neural network for the input \(x_i\), parameterized by $\phi$, which encapsulates the trainable parameters of the network (including weights $W$ and biases $b$). The term $\|f(x_i; \phi) - \hat{x}_i\|^2$ calculates the squared Euclidean norm, summing the squared differences between elements of the predicted output and target data. This norm is then averaged across all $N$ data points in the dataset to compute the mean squared error.
Training an \texttt{EDNN} involves iteratively optimizing the network's parameters ($\phi$) to reduce the loss function. This optimization process can be expressed as:

\begin{equation}
\label{Eq:update_step}
\phi^{(\text{new})} = \text{Optimize} \left( \phi^{(\text{old})}, \nabla \text{Loss} \right)
\end{equation}

Equation~\ref{Eq:update_step} outlines an iterative optimization strategy aimed at enhancing the network's capacity to accurately reconstruct the input data. During each iteration, the parameters of the network are updated ($\phi^{(\text{new})}$) based on the gradient of the loss function ($\nabla \text{Loss}$) and the previous parameter values ($\phi^{(\text{old})}$). The choice of the optimization strategy can vary, encompassing algorithms such as Gradient Descent, Adam, etc~\cite{goodfellow2016deep}. Through this iterative process, the network fine-tunes its parameters to minimize the reconstruction error or loss, thereby improving the fidelity of the data reconstruction and enhancing the model's overall performance on unseen data.

\subsection{F1-score}
\label{sec:F1}

Unlike accuracy, which can be misleading if a model performs well only on one class, the F1-score helps ensure that the model’s performance is reliable across all classes~\cite{10.1007/11941439_114}. It is crucial to report F1-score in various tasks, such as medical classification, because it provides a balanced measure of precision and recall. The F1-score is defined as:
\begin{equation}
\label{eq:f1}
F1 = 2 \cdot \frac{\text{precision} \cdot \text{recall}}{\text{precision} + \text{recall}}
\end{equation}
In Equation ~\ref{eq:f1}, precision is the proportion of true positive predictions among all positive predictions, while recall measures the proportion of actual positives that are correctly identified. This balance is essential when both false positives and false negatives carry significant consequences~\cite{Taha2015}. For instance, in medical classification tasks, a false positive might lead to unnecessary treatments or interventions, while a false negative could result in missed diagnoses and lack of essential treatment.

\section{The P2W Approach}
\label{sec:P2W}
In this section, we describe our unconventional transfer learning approach in detail. The approach is based on the following assumptions. First, we assume that we only have a small dataset, $D_{\text{small}}$, which does not contain sufficient data samples to train an ML model from scratch and to achieve high accuracy. Also, we assume that we have an embedded device with a \emph{target SoC} containing a programmable processor core which runs a \emph{target ML model} having the knowledge we want to transfer. We can neither reprogram the target SoC nor look at the target ML model inside. The publicly available information we have about the target model is its type (MLP or CNN) and its topology. For example, there are many ARM Cortex-M/A based SoCs, known for their affordability and low power consumption, making them ideal for embedded machine learning applications~\cite{noauthor_plumerai_nodate, electronics11162545}. In addition, manufacturers of embedded devices typically use ML models inside such SoCs, taken from publicly available libraries like TinyML~\cite{warden2019tinyml}, STM32 model zoo~\cite{noauthor_stmicroelectronics_2023}, and others. However, the manufacturers often use proprietary training datasets and training approaches to fine-tune these models in order to obtain a specific model which is proprietary as well. 

Second, we assume that we can acquire a clone of the aforementioned target SoC without any ML model inside, and we can easily (re-)program this \emph{clone SoC} and experiment with it. This is feasible because well-established SoC manufacturers provide affordable and easy-to-use HW/SW development kits for their popular SoCs used in many embedded devices. For example, Microchip, STMicro, and others sell evaluation and prototyping boards with SoCs based on ARM Cortex-M/A series of processor cores~\cite{noauthor_evaluation_nodate,noauthor_stm32_nodate}.

\begin{figure*}[!t]
\centerline{\includegraphics[width=1\columnwidth]{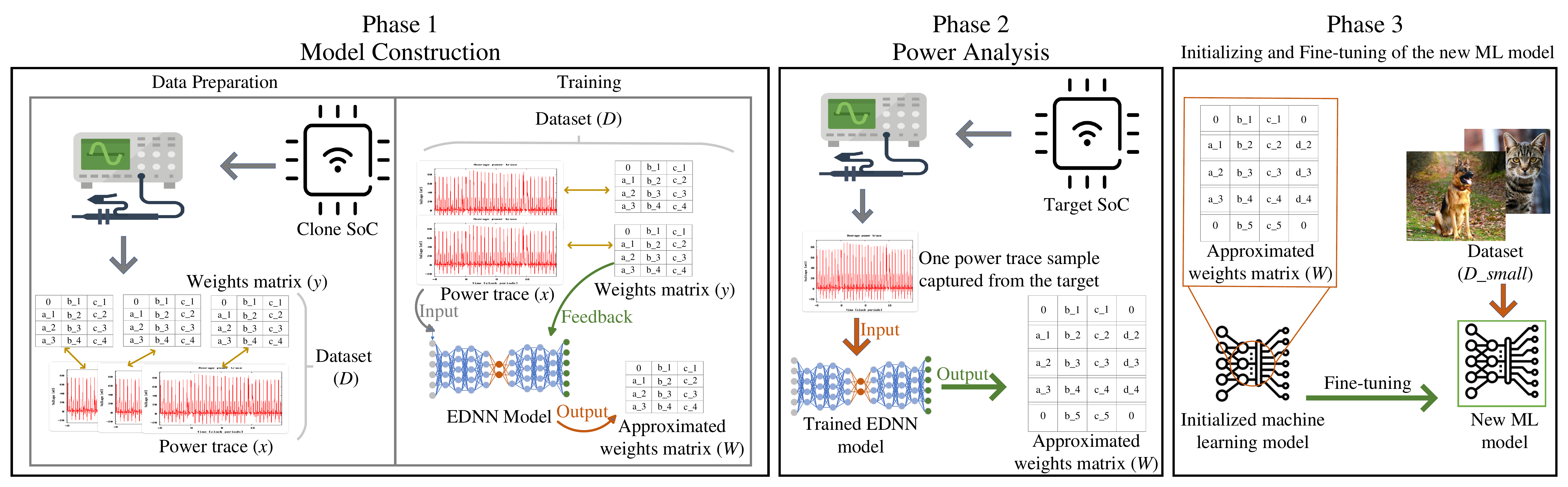}}
    \caption{High-level overview of the phases in the P2W transfer learning approach, including \texttt{EDNN} Model Construction, EDNN-based Power Analysis, and Initializing and Fine-tuning of the new ML model.}
    \label{fig:Introduction_for_the_Template_attack}
\end{figure*}

Considering the above assumptions, Figure~\ref{fig:Introduction_for_the_Template_attack} illustrates the overall workflow of our P2W approach. The main goal of P2W is to transfer knowledge from a well-trained target ML model (MLP or CNN), running inside a target SoC, to a new ML model in order to establish a basis for transfer learning, and then to fine-tune this new model for another related task using the small dataset $D_{\text{small}}$. We do this transfer by analyzing power traces, obtained from the target SoC running the well-trained ML model, with the help of an \texttt{EDNN}. Our P2W approach consists of three phases.

In \textsc{Phase 1}, described in detail in Section~\ref{sec:DataPrep}, we construct the \texttt{EDNN} which tries to learn the relationship between the power consumption of the aforementioned target SoC and the weights of the target ML model inside. Since, we can neither reprogram the target SoC nor look at the target ML model inside, we use the aforementioned clone SoC and the publicly available information about the model type and topology to perform the \texttt{EDNN} construction in two steps called \texttt{Data Preparation} and \texttt{Training}. In the \texttt{Data Preparation} step , we obtain a training dataset $D = \{d_1, d_2, ..., d_{|D|}\}$ for the \texttt{EDNN}. Every $d_i \in D$ is a pair $ d_i= (x_i,y_i)$ where $x_i$ is the power consumption trace captured when the clone SoC runs an inference task using a \emph{surrogate ML model} with a set of weights $y_i$. This surrogate model is of the same type and has the same topology as the target ML model. By reprogramming the clone SoC multiple times to run the surrogate ML model with different sets of weights, we obtain dataset $D$. It is used in the \texttt{Training} step where a standard supervised learning method is applied to train the \texttt{EDNN}.

In \textsc{Phase 2}, called \texttt{EDNN-based Power Analysis}, we use the trained \texttt{EDNN} to extract knowledge from the target SoC. To achieve this goal, only one power trace $x$, captured from the target SoC running the target ML model, is fed into the trained \texttt{EDNN} constructed in \textsc{Phase 1}. The \texttt{EDNN} translates power trace $x$ into weights matrix $W$, whose elements are an approximation of the weights and biases of the target ML model. Our power analysis is inspired by the general ideas and principles briefly introduced in Section~\ref{sec:PowerAnalBackground}.

Finally, in \textsc{Phase 3}, called \texttt{Initializing and Fine-tuning}, we train a new ML model using the aforementioned small dataset $D_{\text{small}}$ and the approximated weights matrix $W$ obtained in \textsc{Phase 2}. First, we initialize the new ML model with $W$, and then we further train/fine-tune this model using $D_{\text{small}}$. The fine-tuning process involves optimizing the weights and biases to adapt the new ML model to perform a specific inference task.

In the rest of this section, we explain \textsc{Phase 1} and \textsc{Phase 3} of our P2W approach in more detail.

\subsection{Phase 1: \texttt{EDNN} Model Construction}
\label{sec:DataPrep}

As previously mentioned, \textsc{Phase 1} consists of two steps. In the first step, \texttt{Data Preparation}, the goal is to prepare dataset $D$ for training the \texttt{EDNN} model. To create this dataset, we deploy the surrogate ML model on the clone SoC and enable it to perform inference. However, the surrogate ML model must have been trained to perform an inference task similar to the task performed by the target ML model. For example, if we want to transfer the knowledge of a target ML model performing a medical image classification task, then we have to deploy a surrogate ML model on the clone SoC that has been trained to perform similar tasks such as object image classification, animal image classification, etc.

The next action in step \texttt{Data Preparation} involves inputting a single random data sample to be processed by the clone SoC running the surrogate ML model, and capturing the SoC's power consumption from the start to the end of the inference process for this random data sample, thereby obtaining the corresponding power trace. It is important to note that the input data sample selected randomly must remain the same throughout the entire process, in both \textsc{Phase 1} and \textsc{Phase 2}. The sampling rate used to capture the power trace depends on the SoC's clock frequency. A suitable sampling rate can be determined experimentally for each case. Our studies show that it is better to keep the sampling rate not lower than $1/3$ of the SoC's clock frequency. However, training the \texttt{EDNN} with long and high resolution power traces in dataset $D$ may increase the computational complexity. To address this issue, we reduce the length of the power traces using a method called Principal Component Analysis (PCA)~\cite{tharwat2016principal}. The primary advantage of PCA is its ability to convert a large set of correlated variables, such as millions of samples in the power trace, into a smaller set of uncorrelated variables known as principal components. This process effectively retains the most significant features of the original dataset, ensuring that essential characteristics of the power trace are preserved, thereby maintaining the integrity of the signal for subsequent analysis.

 The surrogate ML model has to be deployed on the clone SoC multiple times in order to obtain dataset $D$. Each time we deploy and run the surrogate ML model, we have to use the same randomly selected input data sample and a different set of weights $y_i$ in order to capture a new power trace $x_i$ during the model inference.

After dataset $D$ is created, the \texttt{Training} step of \textsc{Phase 1} commences. As illustrated in Figure \ref{fig:Introduction_for_the_Template_attack}, we use $D$ to train an \texttt{EDNN} model including a convolutional encoder that receives data as input and a convolutional decoder that transforms the encoder's output. The topology of the \texttt{EDNN} depends on the accuracy which we want to achieve, the quality of the captured data in dataset $D$, and the training method which we use to train the \texttt{EDNN}. Based on dataset $D$ containing pairs $d_i = (x_i, y_i)$, we use the standard supervised learning process, described in Section~\ref{sec:AutoEncBackground}, to train the \texttt{EDNN}. In this context, $x_i$ represents $X$ in Equation \ref{Eq:loss_function}, while $y_i$ represents $\hat{X}$ in the same equation.

\begin{algorithm}[!t]
\caption{Dataset Preparation and \texttt{EDNN} training}\label{alg:dataset_preparation}
\DontPrintSemicolon
\SetAlgoLined
\Input{$S = \{s_1, ..., s_{|S|}\}, r, surML, genML, \theta$}
\Output{$genML$}
\BlankLine
$acc = 0$; $k = 1$; $Sub \gets \emptyset$;\\
\While{$acc < \theta \land k \leq |S|$}{
$Sub \gets Sub + s_k$; \\
\For{$s \in Sub$}{
        \For{$i  = 1~to~r$}{
            $surML \gets$ \texttt{Train}($surML, s$);\\ 
            Deploy $surML$ onto clone SoC;\\ 
            $x \gets$ Capture power during inference;\\ \bluecomment{\tcp{Generating the weights matrix}}
            $y \gets$ \texttt{Coefficients\_to\_Matrix($surML$)};\\
            $D \gets D + (x, y)$;\\
        }  
} 
$(acc, genML) \gets$ \texttt{Train}($genML, D$);\\
$k = k + 1$;\\
}
\KwRet{genML};\\

\BlankLine

\SetKwProg{Fn}{Function}{:}{}
\Fn{Coefficients\_to\_Matrix(surML)}{
    $y \gets \emptyset$; $j=0$;\\
    \For{$l \in surML.Layers$}{
        \For{$n \in l$}{
            $z=0$;\\ 
            \For{$c \in n$}{
               $y[j][z] = c$;\\
               $z = z + 1$;
            }
            $j = j + 1$;
        }
    }

  \KwRet{$y$};\\
}

\end{algorithm}

Algorithm~\ref{alg:dataset_preparation} details the \texttt{Data Preparation} step (lines 5-15) for creating dataset $D$ and the \texttt{Training} step (line 16) to train the \texttt{EDNN} with the goal of achieving an accuracy of $\theta$ (line 4). This algorithm takes set $S = \{s_1, \ldots, s_{|S|}\}$, a collection of available datasets needed to train surrogate ML model $surML$, and $r$, which indicates the number of times the training process is repeated for the surrogate ML model with each dataset $s_i \in S$. The two other inputs of the algorithm, namely $surML$ and $genML$, represent the surrogate model and the \texttt{EDNN} model, respectively, both initialized with some initial weights. The last input is the required accuracy $\theta$ for the \texttt{EDNN}. The output of Algorithm~\ref{alg:dataset_preparation} is the trained \texttt{EDNN} $genML$.

In line 3 of Algorithm~\ref{alg:dataset_preparation}, three variables are initialized: $acc$, the current accuracy of $genML$; $k$, a counter keeping track of the involved datasets used to train $surML$; and $Sub$, a subset of $S$ initialized to an empty set $\emptyset$.

In lines 4-18, a while loop iterates until the current accuracy of $genML$ exceeds $\theta$ or $k$ exceeds $|S|$, indicating that there are no more datasets in $S$ to continue the training with. In line 5, one dataset $s_k \in S$ is added to $Sub$. Subsequently, from lines 6 to 15, the algorithm repeats the training of $surML$ for each dataset $s \in Sub$, thereby expanding dataset $D$ for $r$ times. In line 8, the surrogate ML model $surML$ is trained with dataset $s$. In line 9, the trained model is deployed to the clone SoC, and after deployment, the power trace $x$ of the clone SoC is captured during inference (line 10). In line 12, all coefficients of $surML$ are put in the weights matrix $y$ with the help of function $Coefficients\_to\_Matrix()$. Subsequently, in line 13, the pair $(x, y)$ containing the power trace $x$ and the corresponding weights matrix $y$ is added to dataset $D$. Finally, in line 16, the $genML$ is trained with dataset $D$, and the current accuracy of $genML$ is saved in $acc$.

\begin{figure}[!t] \centerline{\includegraphics[width=1\columnwidth]{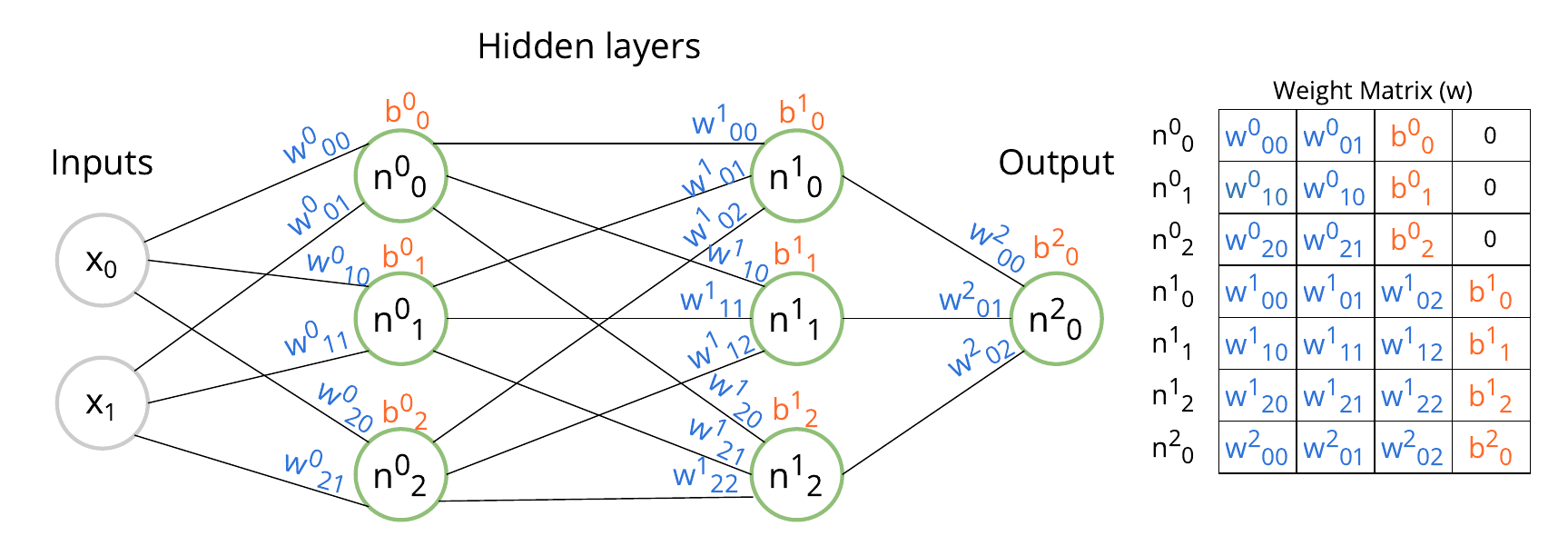}}
  \caption{An example of a weights matrix construction.}
\label{fig:Weight_and_Bias_to_Matrix}
\end{figure}

The behavior of the function $Coefficients\_to\_Matrix()$, used in line 12, is detailed in lines 20-32. This function takes a surrogate model $surML$ as input and returns a matrix $y$ containing all coefficients of $surML$. Within this function, lines 22-31 contain three nested loops. The first loop iterates over each layer $l$ of $surML$. The second loop iterates over each filter/neuron $n$ in layer $l$. The innermost loop iterates over each coefficient $c$ in filter/neuron $n$. In line 26, each coefficient $c$ is added to the weights matrix $y$. An example of the output matrix $y$ from the function $Coefficients\_to\_Matrix()$ applied to an MLP model is shown in Figure \ref{fig:Weight_and_Bias_to_Matrix}. The MLP model consists of four layers, where the first layer contains 2 inputs, both the second and third layers consist of 3 neurons each, and the last one is the output. In the output weights matrix $y$, each row contains the weights and bias from a specific filter/neuron $n^i_j$  where $i$ is the index of the layer and $j$ is the index of the node within that layer. For example, $n^0_2$ represents neuron number two in layer 0, and $w^0_{21}$ represents the coefficient number 1 of neuron $n^0_2$.

\subsection{Phase 3: Initialization and Fine-tuning}

In \textsc{Phase 3}, we aim at training a new ML model using transfer learning. The first step involves initializing the new ML model with the approximated weights matrix $W$, which is generated by the \texttt{EDNN} in \textsc{Phase 2}. Since we keep the topology of the surrogate ML model, the target ML model, and the new ML model the same across all three phases, the shape of the approximated weights matrix $W$ matches the shape of the new ML model's weights matrix. This means that every weight and bias in the new ML model has a corresponding value in the approximated weights matrix $W$ and this value is used for initialization. Such initialization with corresponding values taken from $W$ infuses the new ML model with pre-established knowledge acquired from the target ML model inside the target SoC. This creates a foundation for the new ML model to achieve levels of accuracy comparable with the target ML model.

Following the initialization step, the new ML model is fine-tuned with dataset $D_{\text{small}}$ to obtain the final model as follows:
\begin{equation}
newML_{\text{final}} = \text{Train}(newML_{\text{init}(W)}, D_{\text{small}})
\end{equation}
where \texttt{Train} can be any supervised learning approach suitable to train and optimize the initialized new ML model ($newML_{\text{init}}$) utilizing dataset $D_{\text{small}}$. 
Recall that  $D_{\text{small}}$ does not contain a sufficient number of data samples to successfully train a randomly initialized ML model from scratch and to achieve high accuracy. However, the initialized new ML model with $W$ has a great potential to be trained for its task using this small dataset $D_{\text{small}}$ and achieve high accuracy because a significant amount of knowledge has been transferred from the target ML model through $W$. Finally, strategies such as adaptive learning rate adjustments and early stopping can be employed during this training to fine-tune the model’s parameters faster.

\section{Evaluation}
\label{sec:Evaluation}
In this section, we evaluate our P2W approach. The evaluation focuses particularly on two core aspects: the accuracy of the \texttt{EDNN} models obtained in \textsc{Phase 1} and used in \textsc{Phase 2},  and the overall performance of the new ML models obtained after the fine-tuning in \textsc{Phase 3}. First, we discuss our experimental setup concerning the surrogate ML models (\textsc{Phase 1}), the \texttt{EDNN} models (\textsc{Phase 1} and \textsc{Phase 2}), and the datasets $D_{\text{small}}$ (\textsc{Phase 3}) in Section~\ref{sec:SurrMLTop}, Section~\ref{sec:AutoEncTop}, and Section~\ref{sec:D_small}, respectively. This is followed by the evaluation of the P2W approach in Section~\ref{sec:AutoEncEval},  Section~\ref{sec:NewMLEval}, and Section~\ref{sec:P2WEvalImbalanced}. 

\subsection{Surrogate ML Models and Power Trace Capturing}
\label{sec:SurrMLTop}

In our experiments, we utilize two surrogate ML models, called Model 1 and Model 2, each given as input $surML$ to Algorithm~\ref{alg:dataset_preparation}. These models are two different fully connected neural networks, i.e., MLP networks with different topologies, that are trained in Line 8 of Algorithm~\ref{alg:dataset_preparation} to perform different classification tasks. 

\textbf{Model 1:} This surrogate ML model is trained to perform binary classification using the ECG and Heart Disease datasets described in Table~\ref{tab:datasets}. The model has one input layer followed by four fully connected hidden layers. The first hidden layer consists of 128 neurons and employs the ReLU activation function. The subsequent three hidden layers have 64 neurons each. Also, the model has an output layer with a single neuron utilizing the Sigmoid activation function.

\textbf{Model 2:} This surrogate ML model is trained to perform ternary classification using the Blood Pressure dataset described in Table~\ref{tab:datasets}. The model is a modified version of Model 1 where the number of neurons in the second hidden layer is increased to 128 and the output layer contains three neurons corresponding to the three classes.
  
In Table~\ref{tab:datasets}, each dataset is characterized by four main attributes. The first attribute is \texttt{Size}, representing the number of data samples in each dataset. The second attribute is \texttt{Features}, indicating the length of each sample in bytes. The third attribute is \texttt{Classes}, denoting the number of classes associated with each dataset. The last attribute is \texttt{Chunks}, showing the number of chunks, created from each dataset. In addition, Table~\ref{tab:datasets} shows that the datasets are grouped based on their number of classes, thereby forming two groups, where one group consists of datasets associated with 2 classes and the other with 3 classes.

\begin{table}[!t]
\centering
\caption{Summary of the datasets utilized in the training of the surrogate ML models in \textsc{Phase 1}.}

\label{tab:datasets}
\resizebox{\columnwidth}{!}{%
\begin{tabular}{lccccc}
\hline
 \textbf{Dataset} & \textbf{Size} & \textbf{Features} & \textbf{Classes} & \textbf{Chunks} & \textbf{Ref}\\ 
\hline
 ECG              & 109k            & 188          & 2     &100 &\cite{Kachuee_2018}   \\
 Heart Disease     & 319k          & 16            & 2  &310 &\cite{georgy_blood_nodate}\\
 \hline
 Blood Pressure     & 1k          & 50            & 3  &10 &\cite{pavan_bodanki_blood_nodate}\\
 \hline

\end{tabular}
}
\end{table}

The collection of all chunks belonging to a group of datasets is given as input $S = \{s_1, ..., s_{|S|}\}$ to Algorithm~\ref{alg:dataset_preparation}. Every $s_i \in S$ is a chunk which is used to train the surrogate ML models separately as shown in Line 8 of the algorithm. Moreover, input $r$ of the algorithm is set to 30. Thus, we repeat multiple times the training of any surrogate model $surML$ with each chunk of data $s$ (Lines 7-8). During each training round, to address the problem of overfitting, a dropout strategy is integrated with a rate of 0.3 applied after each fully connected layer. This strategy is critical for enhancing the model’s generalization capabilities and maintaining consistent performance across different datasets. 

In Lines 9-10 of Algorithm~\ref{alg:dataset_preparation}, to capture power traces during the inference of trained surrogate model $surML$ deployed on the clone SoC, we use the ChipWhisperer platform~\cite{o2014chipwhisperer}, with the ChipWhisperer-Huskey as the power capture device and the CW313 SAM4S board as the clone SoC. The board is equipped with a 32-bit ARM® Cortex®-M4 RISC processor and allows for a power sampling frequency of 13 MS/s at a resolution of 12 bits per power sample.

Finally, the aforementioned chunk-based repetitive training of surrogate models \textbf{Model 1} and \textbf{Model 2} with capturing of power traces result in one dataset $D$ containing 2000 distinct pairs $(x, y)$ and another one containing 2800 distinct pairs, respectively. Every dataset $D$ is partitioned and used as follows: 75\% used for training the \texttt{EDNN} model, 20\% dedicated to testing it, and 5\% reserved for validation.

\subsection{EDNN Models and Training Parameters} 
\label{sec:AutoEncTop}
We construct and train our \texttt{EDNN} model in Line 16 of Algorithm~\ref{alg:dataset_preparation} using PyTorch \cite{10.5555/3454287.3455008}, primarily thanks to its flexibility and performance. The \texttt{EDNN} topology, given as input $genML$ to Algorithm~\ref{alg:dataset_preparation}, consists of two parts: Encoder and Decoder. Recall that the purpose of the trained \texttt{EDNN} is to translate a one-dimensional (1D) power trace $x$ into a two-dimensional (2D) weights matrix $y$ (Section~\ref{sec:DataPrep}). 

\textbf{Encoder Topology: }
The encoder part is designed to process a 1D input array of size 1024 elements, where the array represents a power trace. The encoding part starts with a fully connected layer containing 256 neurons followed by a sequence of three 1D convolutional layers. Each of these layers utilizes a kernel with a size of 4 and a stride of 1. The first convolutional layer contains 256 filters, focusing on extracting mid-level features from the input. It is succeeded by a layer with 128 filters. The final convolutional layer, equipped with 64 filters, completes the feature extraction process. ReLU is used as the activation function in convolutional layers.

\textbf{Decoder Topology:}
The decoder part takes as an input the features extracted by the encoder part. The decoder part starts with two sequentially arranged 1D transposed convolutional layers, designed to upscale the input feature maps while enhancing spatial details. The first of these layers consists of 256 filters, employs a kernel with a size of 5 and a stride of 1, and incorporates activation function ReLU for non-linearity. The second transposed convolutional layer, employs a kernel with a size of 4 and a stride of 1, has 128 filters, aiming at refining the feature maps. The feature maps are processed by a linear (fully connected) layer whose number of neurons has to match the size of the 2D output weights matrix. This layer is crucial for reshaping the decoder's output to its final dimensions. 

\textbf{Training parameters:}
Using the \texttt{EDNN} model topology described above, two different \texttt{EDNN} models $genML$ are constructed by executing Algorithm~\ref{alg:dataset_preparation} two times, each time with a different surrogate ML model, given as input $surML$, and with an input value of $\theta = 85\%$. The surrogate ML models \textbf{Model 1} and \textbf{Model 2}, described in Section~\ref{sec:SurrMLTop}, are used for this purpose. When Algorithm~\ref{alg:dataset_preparation} is executed with $surML$ = \textbf{Model 1}, the \texttt{EDNN} training performed in Line 16 is carried out for 100 epochs and when $surML$ = \textbf{Model 2} the training is carried out for 130 epochs. A batch size of 100 and a learning rate of 0.001 are employed for all training epochs. We utilise the mean squared error loss function to quantitatively measure the difference between the actual and predicted weights matrices. The Adam optimizer is used during the training thanks to its adaptability and efficiency \cite{Hassan2023}.
To mitigate the risk of overfitting and ensure the model robustness, a dropout mechanism with a rate of 0.5 is applied after each transposed convolutional operation during the training.

\subsection{Balanced Datasets $D_{\text{small}}$}
\label{sec:D_small}

\begin{table}[!t]
\centering
\tiny

\caption{Summary of the small datasets $D_\text{small}$ utilized in \textsc{Phase 3} for fine-tuning.}
\label{tab:small_datasets}
\resizebox{\columnwidth}{!}{%
\begin{tabular}{cccccc}
\hline
\textbf{Dataset}                                                  & \textbf{\begin{tabular}[c]{@{}c@{}}Classes\end{tabular}} & \textbf{Size} & \textbf{Features} & \textbf{\begin{tabular}[c]{@{}c@{}}Acc\end{tabular}} & \textbf{Ref} \\ \hline
\begin{tabular}[c]{@{}c@{}}EEG\end{tabular}   & 2 classes                                                            & 22            & 19                & 31\%                                                                          & \cite{data4010014}             \\ \hline
\begin{tabular}[c]{@{}c@{}}Diabetes\end{tabular}   & 2 classes                                                            & 45            & 9                 & 43\%                                                                          & \cite{Smith1988-ux}           \\ \hline
\begin{tabular}[c]{@{}c@{}}Sleep\end{tabular} & 3 classes                                                            & 150           & 13                & 40\%                                                                          & \cite{laksika_tharmalingam_sleep_nodate}           \\ \hline
\end{tabular}%
}
\end{table}

In this section, we introduce the datasets $D_{\text{small}}$ used for the fine-tuning in \textsc{Phase 3}. Recall that our P2W approach assumes that $D_{\text{small}}$ does
not contain a sufficient number of data samples to successfully train a randomly initialized ML model from scratch and to achieve high accuracy. We confirm this by showing and discussing a set of experiments that evaluate the accuracy of new ML models trained with $D_{\text{small}}$ only.

Table~\ref{tab:small_datasets} refers to three datasets (column \texttt{Ref}), where the EEG and  Diabetes datasets contain data samples belonging to two classes, and the Sleep dataset contains samples belonging to three classes. It should be noted that these original datasets are relatively large and balanced. A balanced dataset is a dataset within which all classes have (almost) the same number of data samples. In order to perform experiments showing the performance of P2W in scenarios where we do not have access to a sufficient number of data samples, we use small batches of data samples taken from these large balanced datasets. Thus, whenever we refer to balanced $D_{\text{small}}$ in our experiments, it means a small balanced batch of samples from these datasets. The number of data samples in the small batches is mentioned in the \texttt{Size} column of Table~\ref{tab:small_datasets}. The \texttt{Features} column indicates the length of each sample in bytes. 

To confirm that the number of data samples in the aforementioned datasets $D_{\text{small}}$ 
is not sufficient to train a new ML model from scratch and achieve high (acceptable) accuracy, 
we take 20 randomly selected and balanced $D_{\text{small}}$ datasets and one \emph{test dataset} from each large dataset (EEG, Diabetes, and Sleep). 
We take randomly initialized ML models that have the same topology as \textbf{Model 1} and \textbf{Model 2}, described in Section~\ref{sec:SurrMLTop}, and train them using the $D_{\text{small}}$ datasets. For example, Model 2 is trained with the 20 randomly selected and balanced $D_{\text{small}}$ datasets from the large Sleep dataset. After each training round, we check the accuracy of the model with the corresponding \emph{test dataset} and take the average over the 20 rounds. The average accuracy is reported in the \texttt{Acc} column of Table~\ref{tab:small_datasets} and is below 43\% for all datasets $D_{\text{small}}$. Such low accuracy clearly indicates that the number of data samples (column~\texttt{Size}) in our datasets $D_{\text{small}}$ is not sufficient to train a new ML model from scratch and achieve acceptable accuracy on unseen data.

\subsection{Accuracy Evaluation of the EDNN Models}
\label{sec:AutoEncEval}
        
To evaluate the quality of the approximated weights matrices generated by our \texttt{EDNN} models, we apply P2W with all three phases discussed in Section~\ref{sec:P2W}, but without fine-tuning in \textsc{Phase 3}. In \textsc{Phase 1}, we utilize two surrogate ML models, \textbf{Model 1} and \textbf{Model 2}, to create two datasets $D$ and to train two separate \texttt{EDNN} models using Algorithm~\ref{alg:dataset_preparation}. All these actions are described in Section~\ref{sec:SurrMLTop} and \ref{sec:AutoEncTop}.

In \textsc{Phase 2}, which is the power analysis in P2W, we use the two \texttt{EDNN} models and three target SoCs. Each target SoC runs a well-trained target ML model, whose knowledge we are interested to transfer to a new ML model using one of the \texttt{EDNNs}. By experimenting with three target ML models, we evaluate the \texttt{EDNNs} in three different scenarios. In the first two scenarios the well-trained target ML models perform EEG binary classification and Diabetes binary classification, respectively. For these two scenarios, the first \texttt{EDNN} is obtained in \textsc{Phase 1} by utilizing surrogate~\textbf{Model 1}. In the third scenario, the well-trained target ML model performs Sleep ternary classification, and the second \texttt{EDNN} is obtained by utilizing surrogate~\textbf{Model 2}.  

Each of the three well-trained target ML models runs on the CW313 SAM4S board which is our target SoC in the three scenarios.  This board is part of the same ChipWhisperer platform, introduced in Section~\ref{sec:SurrMLTop}, because we use this platform to capture power traces from the target SoC. In \textsc{Phase 2}, while each of the target ML models is performing inference, we capture one power trace from the target SoC and feed the power trace to the corresponding \texttt{EDNN}. The \texttt{EDNN} generates an approximated weights matrix for the target model.

\begin{figure}[!t]
    \centering

    \begin{subfigure}[b]{1\textwidth}
        \includegraphics[width=\textwidth]{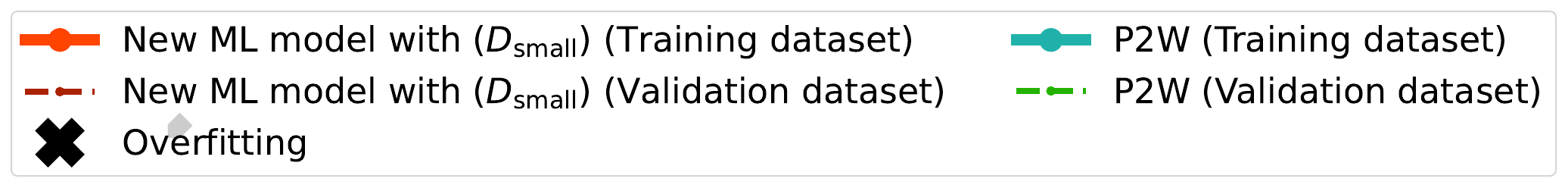}
       
        \label{fig:horizontal}
    \end{subfigure}

    \begin{subfigure}[b]{0.31\columnwidth}
        \includegraphics[width=\textwidth]{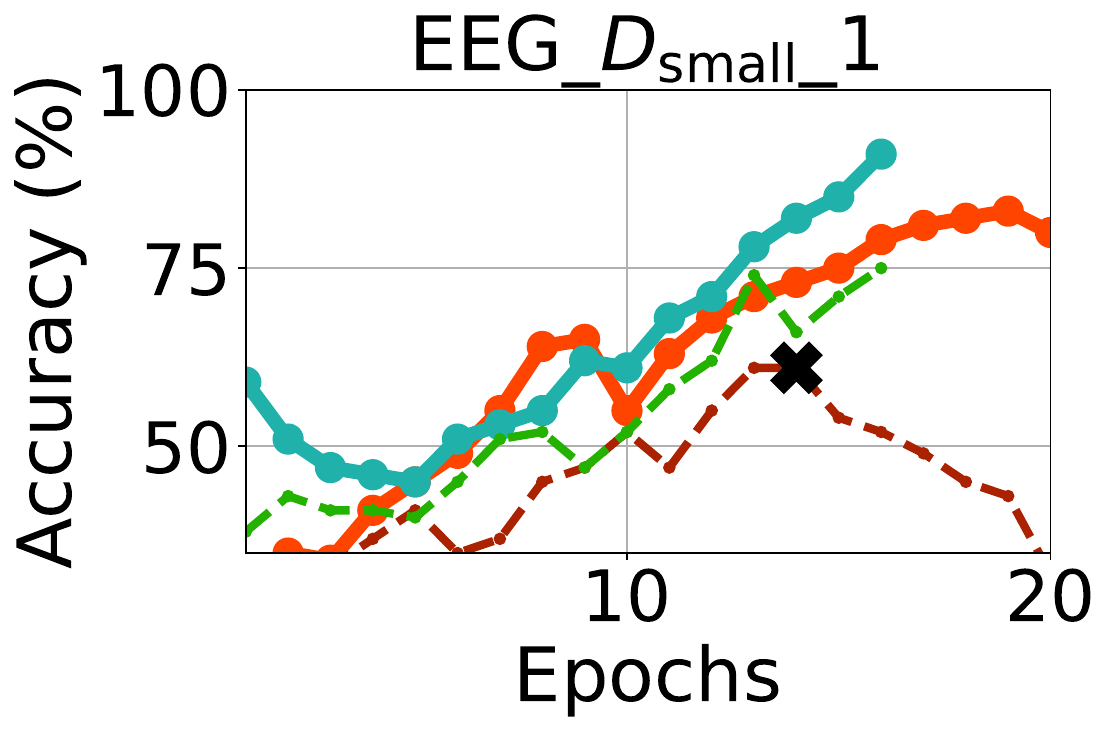}
        \label{fig:EEG_Bach_1}
    \end{subfigure}
    \hfill
    \begin{subfigure}[b]{0.31\columnwidth}
        \includegraphics[width=\textwidth]{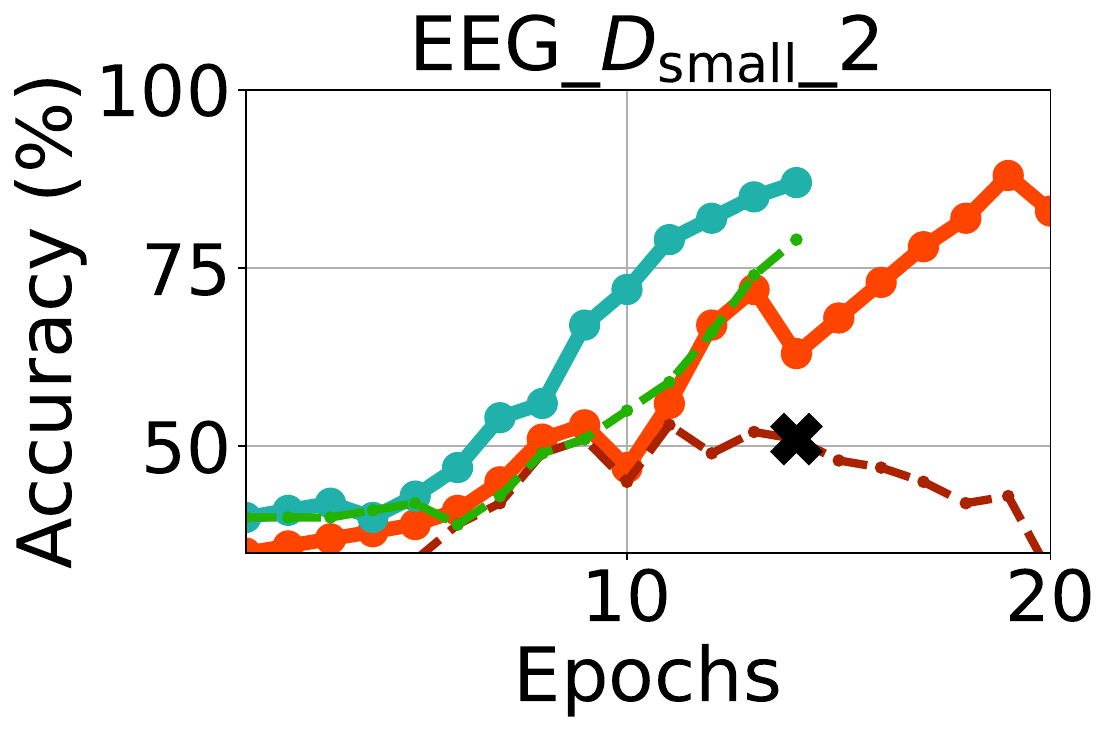}
        \label{fig:EEG_Bach_2}
    \end{subfigure}
    \hfill
    \begin{subfigure}[b]{0.31\columnwidth}
        \includegraphics[width=\textwidth]{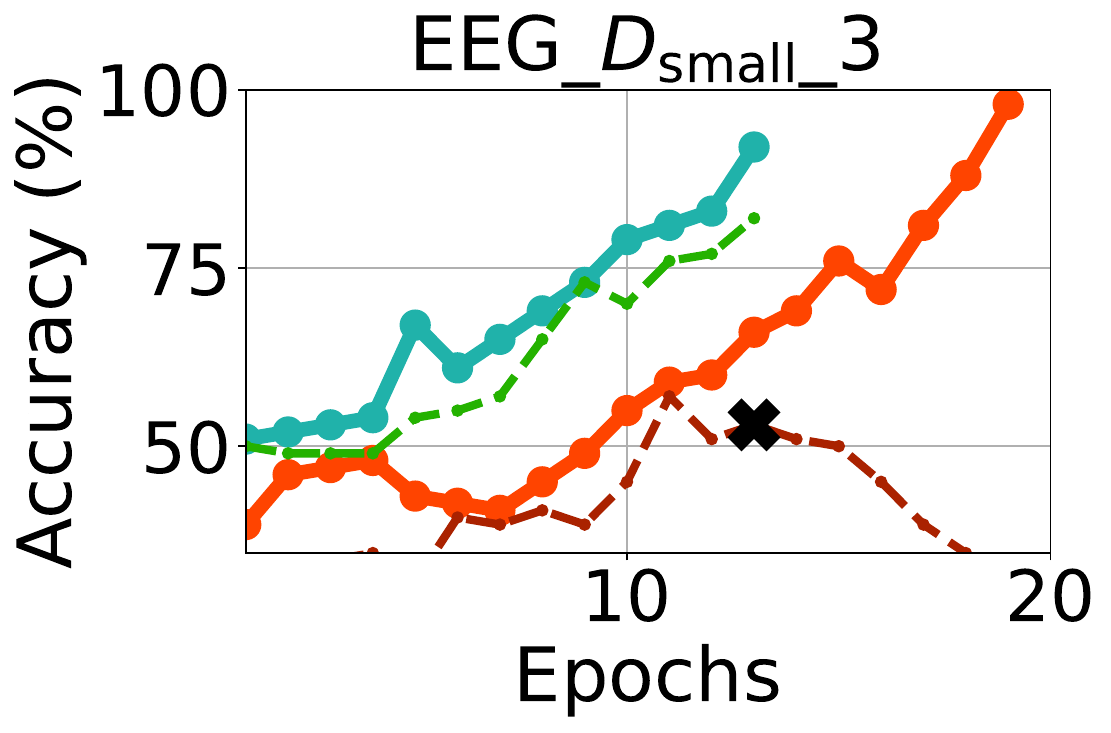}
        \label{fig:EEG_Bach_3}
    \end{subfigure}
    \vspace{-10pt}
    
    \begin{subfigure}[b]{0.31\columnwidth}
        \includegraphics[width=\textwidth]{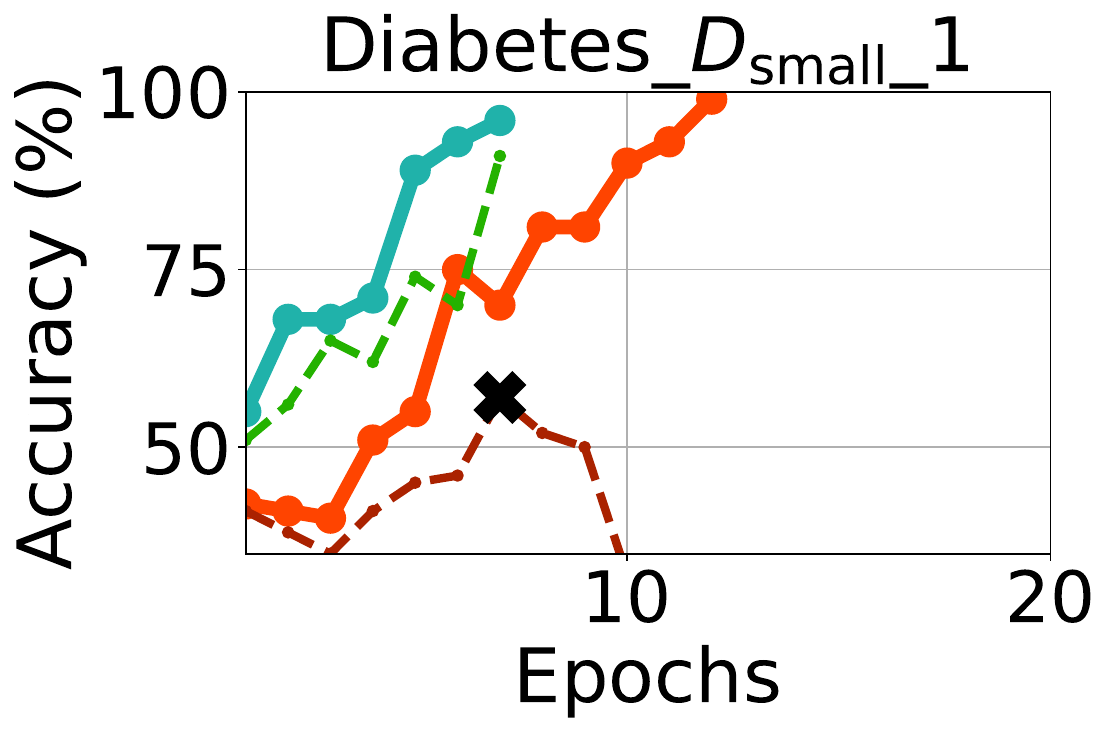}
        \label{fig:Diabetes_Bach_1}
    \end{subfigure}
    \hfill
    \begin{subfigure}[b]{0.31\columnwidth}
        \includegraphics[width=\textwidth]{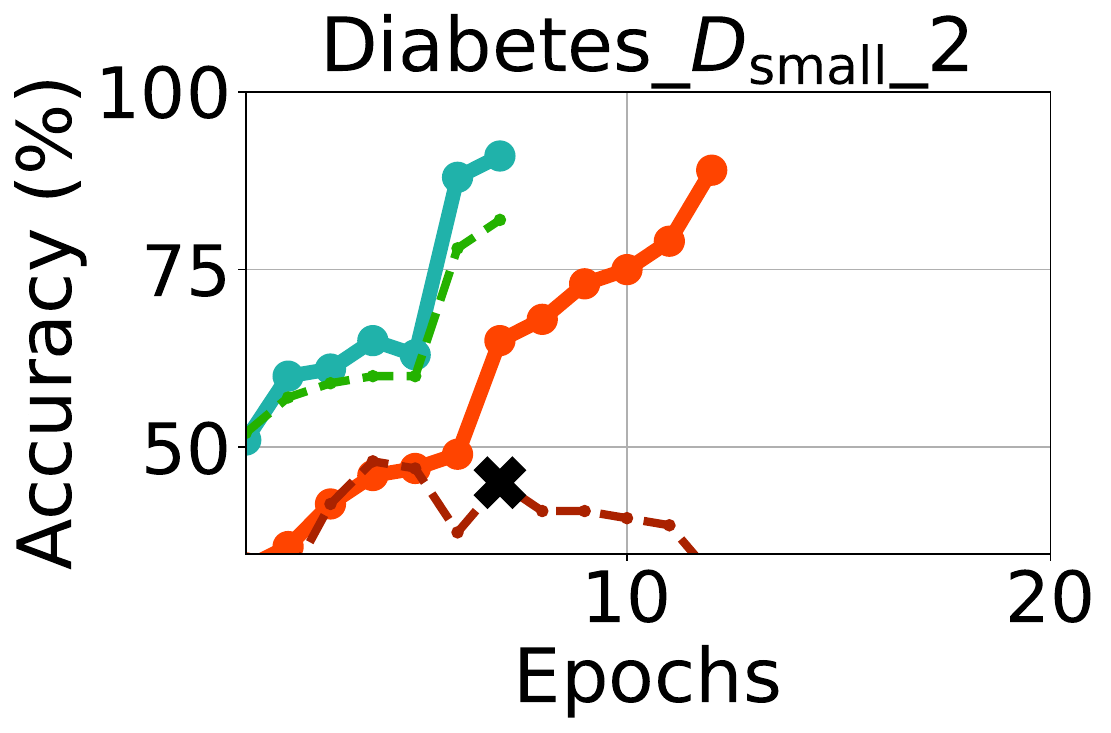}
        \label{fig:Diabetes_Bach_2}
    \end{subfigure}
    \hfill
    \begin{subfigure}[b]{0.31\columnwidth}
        \includegraphics[width=\textwidth]{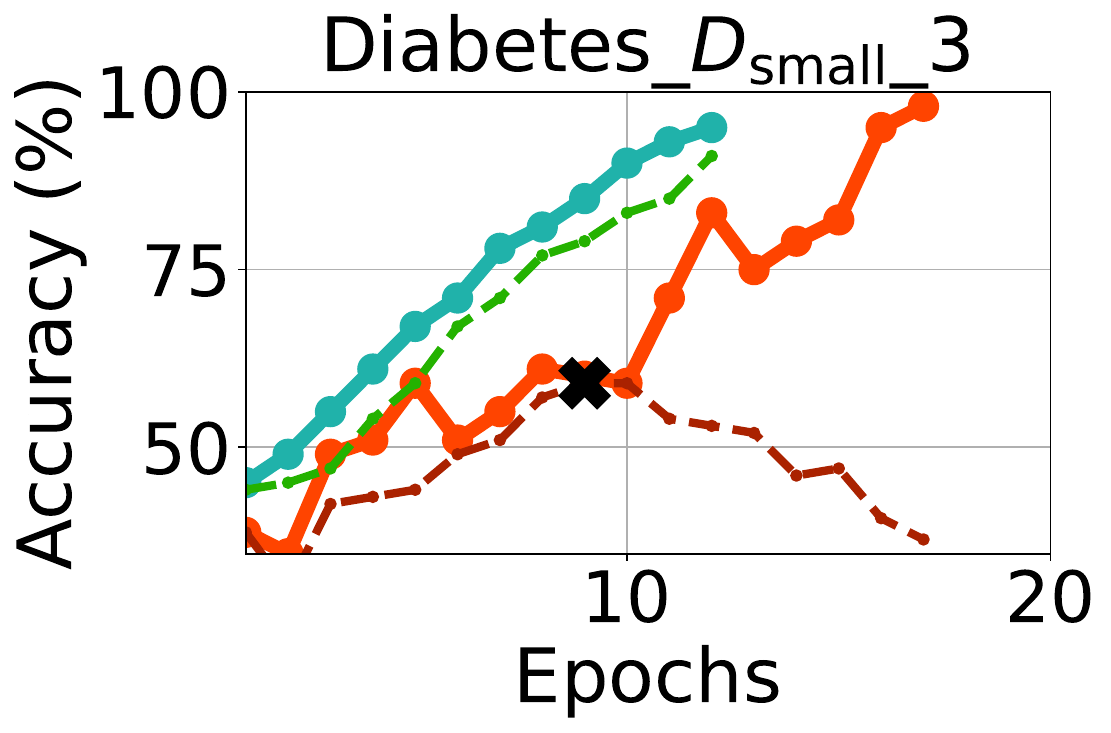}
        \label{fig:Diabetes_Bach_3}
    \end{subfigure}
    \vspace{-10pt}
    
    \begin{subfigure}[b]{0.31\columnwidth}
        \includegraphics[width=\textwidth]{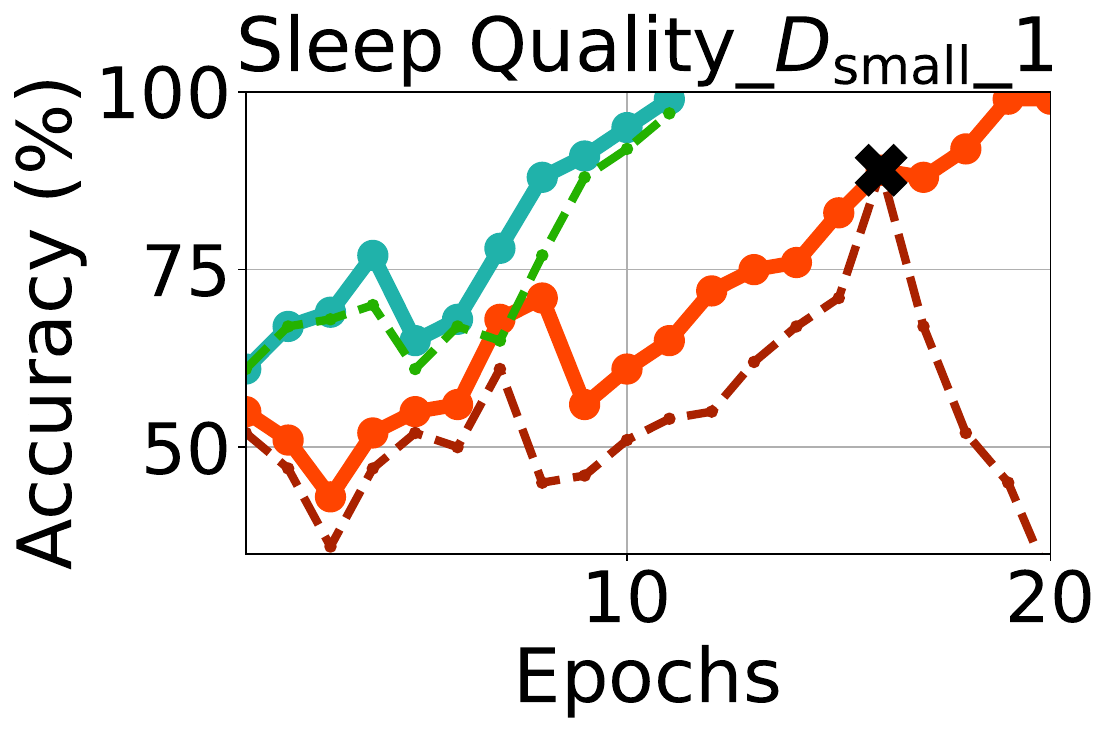}
        \label{fig:Sleep Quality_1}
    \end{subfigure}
    \hfill
    \begin{subfigure}[b]{0.31\columnwidth}
        \includegraphics[width=\textwidth]{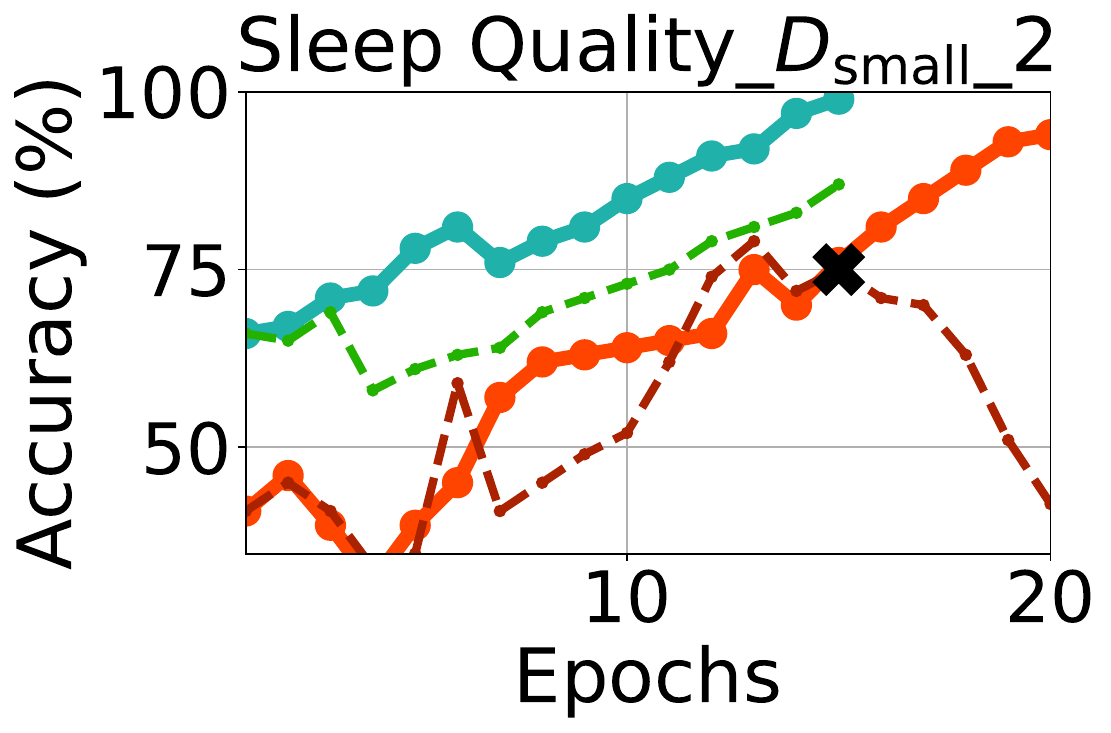}
        \label{fig:Sleep Quality_Bach_2}
    \end{subfigure}
    \hfill
    \begin{subfigure}[b]{0.31\columnwidth}
        \includegraphics[width=\textwidth]{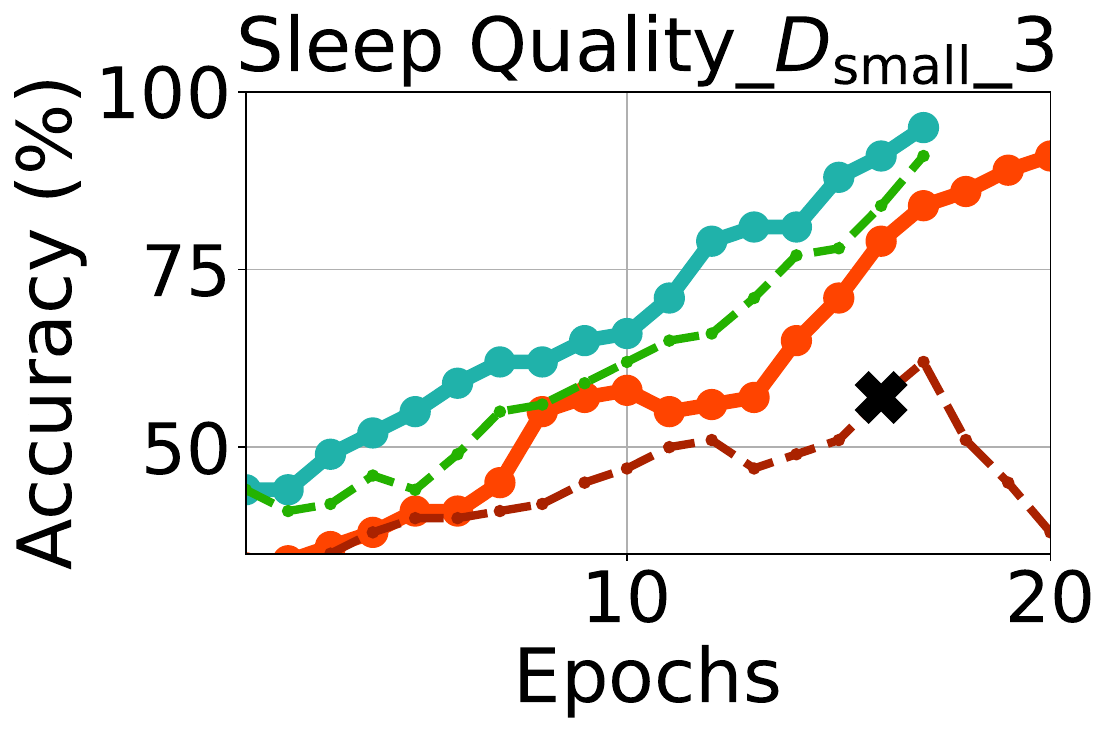}
        \label{fig:Sleep Quality_Bach_3}
    \end{subfigure}
 \vspace{-20pt}
    
    \caption{Accuracy of new ML models trained using only $D_\text{small}$ vs. new ML models obtained using the P2W approach.}
    \label{fig:acuracy_overfitting_12}
\end{figure}

After the aforementioned \textsc{Phase 2}, we obtain three approximated weights matrices (one for every scenario) and we initialize three new ML models with the matrices in \textsc{Phase 3}. In order to evaluate the quality of the approximated weights matrices generated by the \texttt{EDNNs}, we perform a comparison between the accuracy of the target ML models and the new ML models initialized with the matrices. In this way, we evaluate how much knowledge had been transferred from a target ML model to a new ML model with the goal of performing the same classification task as the target model. For this comparison, we test the accuracy of the target and new models by inferring the data samples in the corresponding \emph{test datasets} described in Section~\ref{sec:D_small}. These test datasets are unseen by both the target and new ML models.

Figure~\ref{plo:Compare_generated_original_acc} summarizes the results from our comparison experiment. The horizontal axis shows the three scenarios with the corresponding classification tasks. Each bar represents the accuracy of the models. The gray bars show the average accuracy (column~\texttt{Acc} in Table~\ref{tab:small_datasets}) of the new ML models trained only with balanced $D_{\text{small}}$. The blue bars show the average accuracy of the new ML models initialized with the approximated weights matrix generated by the \texttt{EDNNs} (P2W without fine-tuning). 
The red bars show the accuracy of the target ML models running within the target SoC.

As can be seen in Figure~\ref{plo:Compare_generated_original_acc}, the average accuracy of the new ML models trained only with balanced $D_{\text{small}}$ (gray bars) is too low compared to the target ML models accuracy (red bars). This comparison is also another proof for the fact that datasets $D_{\text{small}}$ are too small and insufficient to train a new ML model from scratch. Now, let us analyze the average accuracy of the new ML models initialized with the approximated weights matrices generated by the \texttt{EDNNs} -- see the blue bars in  Figure~\ref{plo:Compare_generated_original_acc}. These bars show that utilizing weights produced by the \texttt{EDNNs} results in ML models with a notable increase in inference accuracy compared to the gray bars. Although this average accuracy is not as high as the accuracy of the target ML models (the red bars), it indicates that our EDNN-generated weights are non-random and meaningful in the sense that a large amount (more than 60\%) of the knowledge in the target ML models has been transferred successfully to the new ML models via the approximated weights matrices. This fact underscores the efficacy of the \texttt{EDNN} models in generating viable approximated weights matrices that could be used as a good starting point to continue the training and fine-tuning of the new ML model to achieve a higher level of accuracy. This will be demonstrated by the experiments described in the next section.

\subsection{Performance Evaluation of P2W with \\ Balanced Datasets $D_{\text{small}}$}
\label{sec:NewMLEval}

As mentioned in Section~\ref{sec:AutoEncEval}, the initialization of the new ML models with the approximated weights matrices could be a good starting point for further training and fine-tuning in \textsc{Phase 3} with the final goal of obtaining highly accurate new ML models using our proposed P2W transfer learning approach. To continue with the training and fine-tuning of the initialized ML models, we use the small datasets $D_{\text{small}}$ introduced in Section \ref{sec:D_small}. It is important to note that we use these small datasets to highlight the effectiveness of our P2W approach in scenarios with very limited data availability for training.

To visualize the effectiveness of our P2W approach, Figure~\ref{fig:acuracy_overfitting_12} shows some of the obtained training and validation accuracy results for the new ML models trained with P2W and the new ML models trained only with the datasets $D_{\text{small}}$. This figure contains nine plots, each corresponding to an ML model trained once with P2W and another time only with one of the $D_{\text{small}}$ datasets.

For each plot in Figure~\ref{fig:acuracy_overfitting_12}, the horizontal axis represents the number of epochs, while the vertical axis shows the accuracy. The blue-dotted line and the green-dashed line visualize the evolution of the training and validation accuracy of the new ML model trained with P2W, respectively, across the training epochs. Conversely, the orange-dotted line and the red-dashed line visualize the evolution of the training and validation accuracy of the new ML model trained only with $D_{\text{small}}$, respectively, across the training epochs. Additionally, the point at which the new ML model trained only with $D_{\text{small}}$ starts overfitting during training is marked with a black cross.

The results presented in Figure~\ref{fig:acuracy_overfitting_12} show a clear pattern when a randomly initialized ML model is trained only with $D_{\text{small}}$. As the number of epochs increases, (1) the training accuracy (orange-dotted line) continues to improve, but (2) the validation accuracy (red-dashed line) drops at a certain point (black cross). This indicates that the model’s ability to generalize and perform accurate classification on unseen data decreases due to overfitting (black cross). On the other hand, for the models trained with P2W, both training accuracy and validation accuracy keep increasing together, achieving a higher level of accuracy even with a smaller number of epochs.

\begin{figure}[!t]
\centerline{\includegraphics[width=0.8\columnwidth]{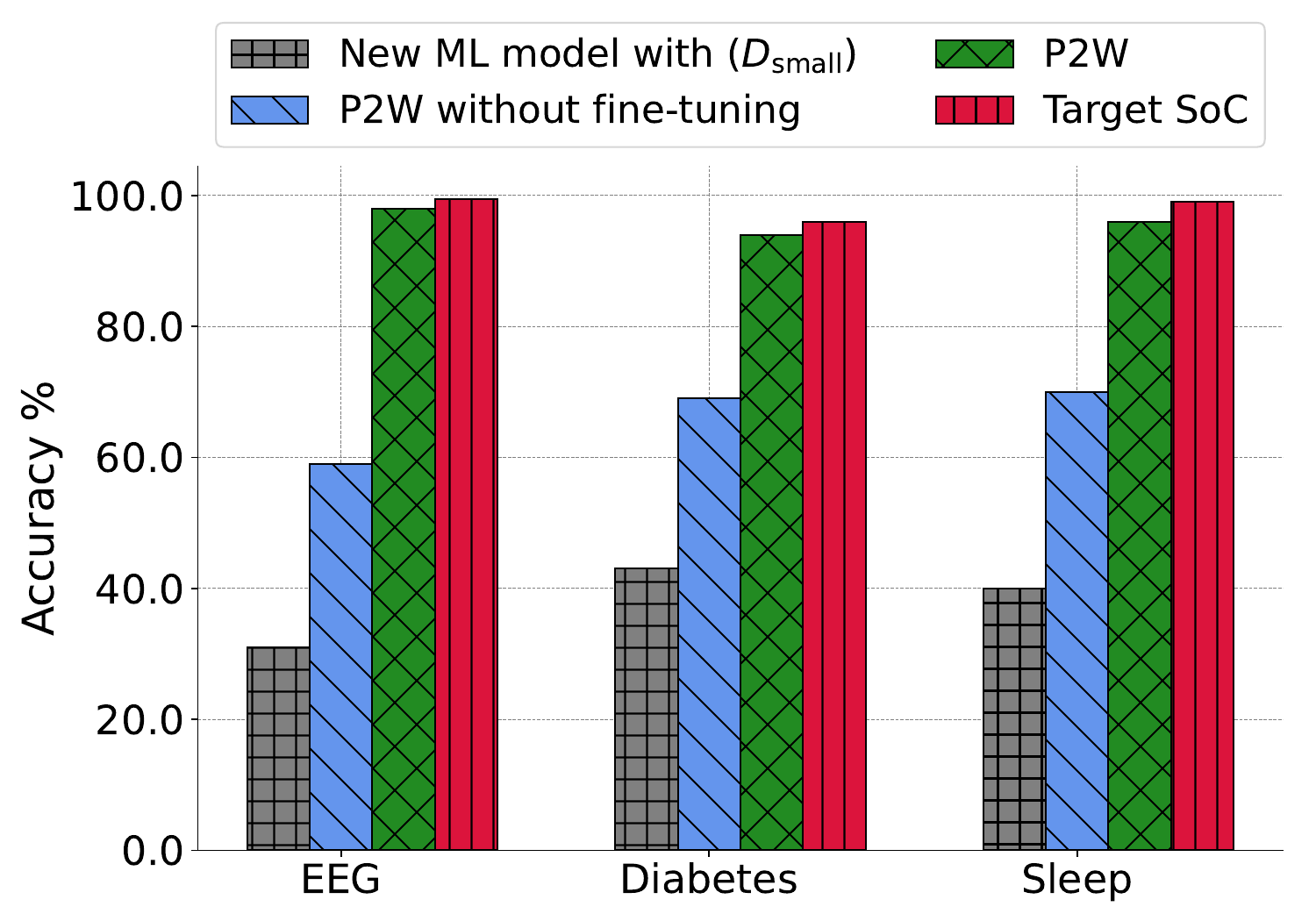}}
\vspace{-10pt}
    \caption{Accuracy of new ML models and target ML models performing EEG, Diabetes, and Sleep classification tasks.}
    \label{plo:Compare_generated_original_acc}
\end{figure}

After training and fine-tuning the initialized ML models with datasets $D_{\text{small}}$, we check the average accuracy of the final new ML models trained with P2W, shown with the green bars in Figure~\ref{plo:Compare_generated_original_acc}, by using the corresponding \emph{test datasets} (Section~\ref{sec:D_small}). Comparing the green bars with the other bars in the figure, we see that new ML models, trained only with $D_{\text{small}}$ (gray bars) or obtained only by initialization with the approximated weights matrices (blue bars), have much lower accuracy compared to the new ML models obtained with our P2W transfer learning approach (green bars). On the other hand, as shown in Figure~\ref{plo:Compare_generated_original_acc}, the accuracy of the new ML models obtained with P2W is still a bit lower than the accuracy of the target ML models (red bars). The reason is twofold: (1) it is practically impossible to transfer 100\% of the knowledge from the target ML models to the new ML models via the approximated weights matrices generated from power traces; (2) the training/fine-tuning of the models in \textsc{Phase 3} is performed with small/limited datasets $D_{\text{small}}$ that are assumed to be the only datasets available to the user of our P2W approach.

Nevertheless, the green bars in Figure~\ref{plo:Compare_generated_original_acc} clearly indicate that new ML models, obtained by P2W, achieve accuracy comparable to the corresponding  target ML models. Moreover, in contrast to the target ML models, these new ML models are fully open and available to any user of P2W for further (re-)use, e.g., deployment of the new ML models on different computing platforms and in different application scenarios, fine-tuning them further to perform different tasks, etc.

\subsection{Performance Evaluation of P2W with Imbalanced Datasets $D_{\text{small}}$ 
} \label{sec:P2WEvalImbalanced}

In Section~\ref{sec:NewMLEval}, we evaluate the performance of our P2W approach in typical scenarios, i.e., scenarios with available \emph{balanced} datasets $D_{\text{small}}$. However, the availability of such balanced datasets may not be always guaranteed due to the limited size of $D_{\text{small}}$. Therefore, in this section, we evaluate the performance of P2W in extreme scenarios, i.e., scenarios with \emph{imbalanced} $D_{\text{small}}$. The goal is to investigate the robustness of P2W to imbalanced training data. An imbalanced dataset is a dataset  within which one or some of the classes have
a much greater number of data samples than the others. Such imbalance can cause biased training of a new ML model, thus unreliable predictive accuracy of the model.

\begin{figure}[!t]
    \centering

    \begin{subfigure}[b]{\textwidth}
        \includegraphics[width=\textwidth]{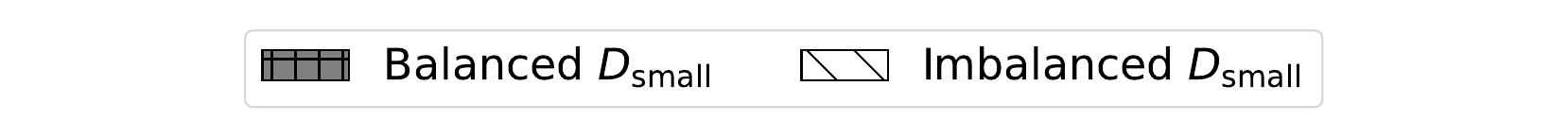}
    \end{subfigure}

    \begin{subfigure}[b]{0.49\columnwidth}
        \includegraphics[width=\textwidth]{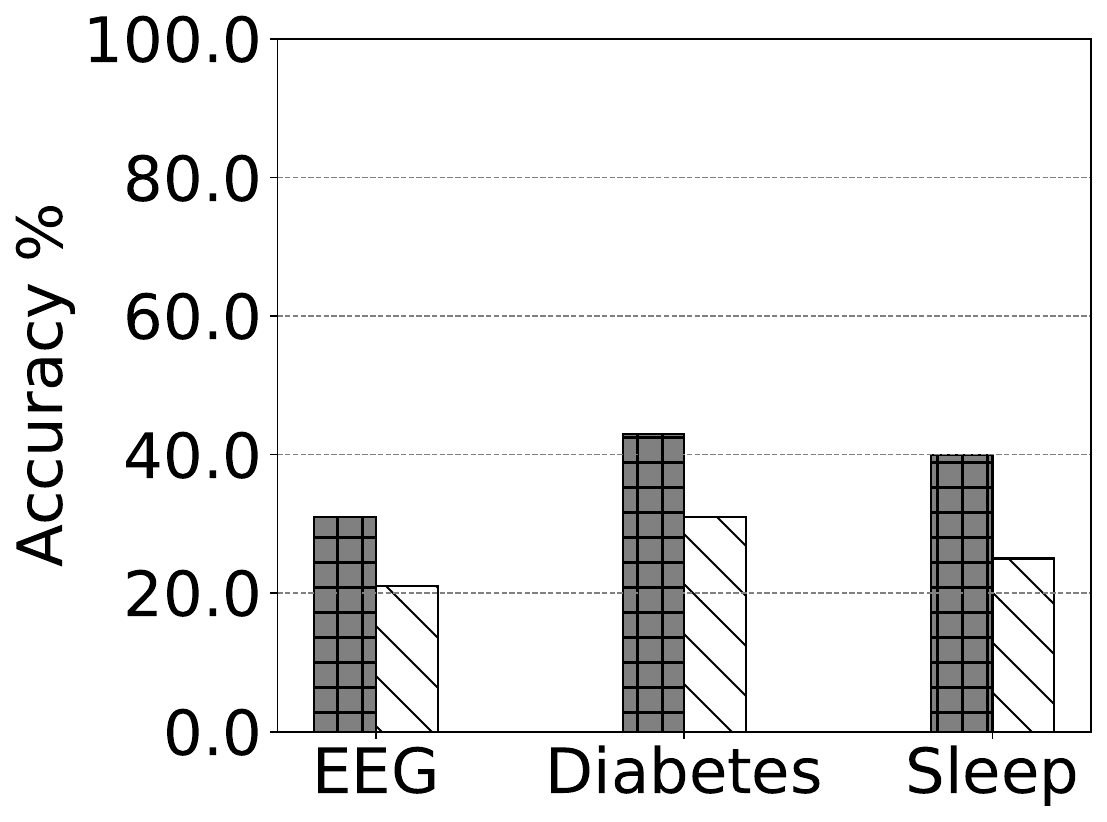}
        \caption{Accuracy without P2W}
        \label{fig:acc}
    \end{subfigure}
    \hfill
    \begin{subfigure}[b]{0.49\columnwidth}
        \includegraphics[width=\textwidth]{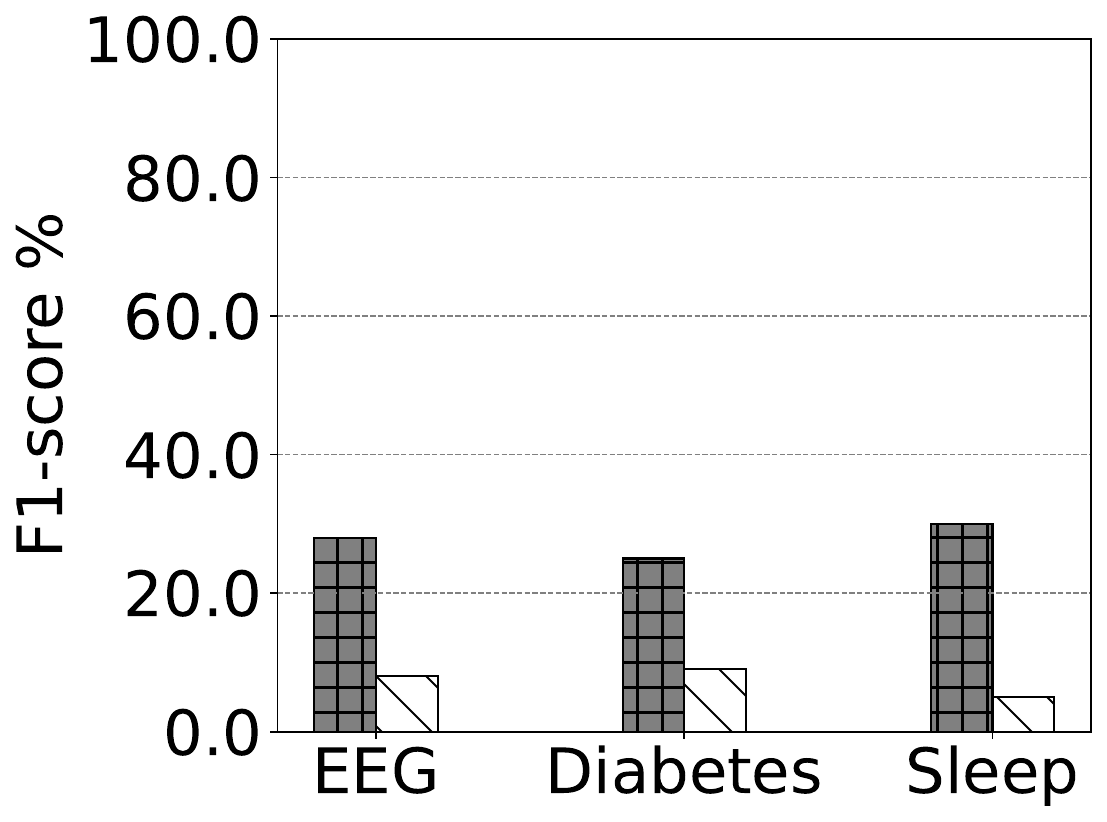}
        \caption{F1-score without P2W}
        \label{fig:f1_score}
    \end{subfigure}
   
\vspace{2pt}

    \begin{subfigure}[b]{\textwidth}
        \includegraphics[width=\textwidth]{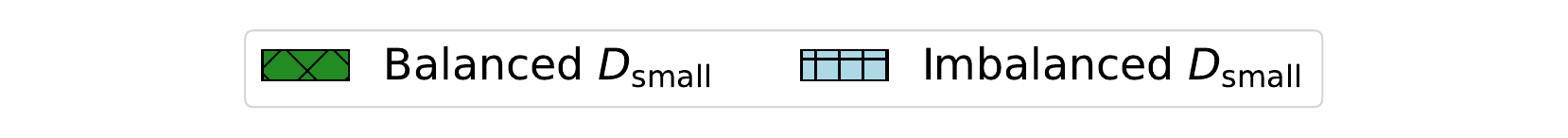}
    \end{subfigure}

    \begin{subfigure}[b]{0.49\columnwidth}
        \includegraphics[width=\textwidth]{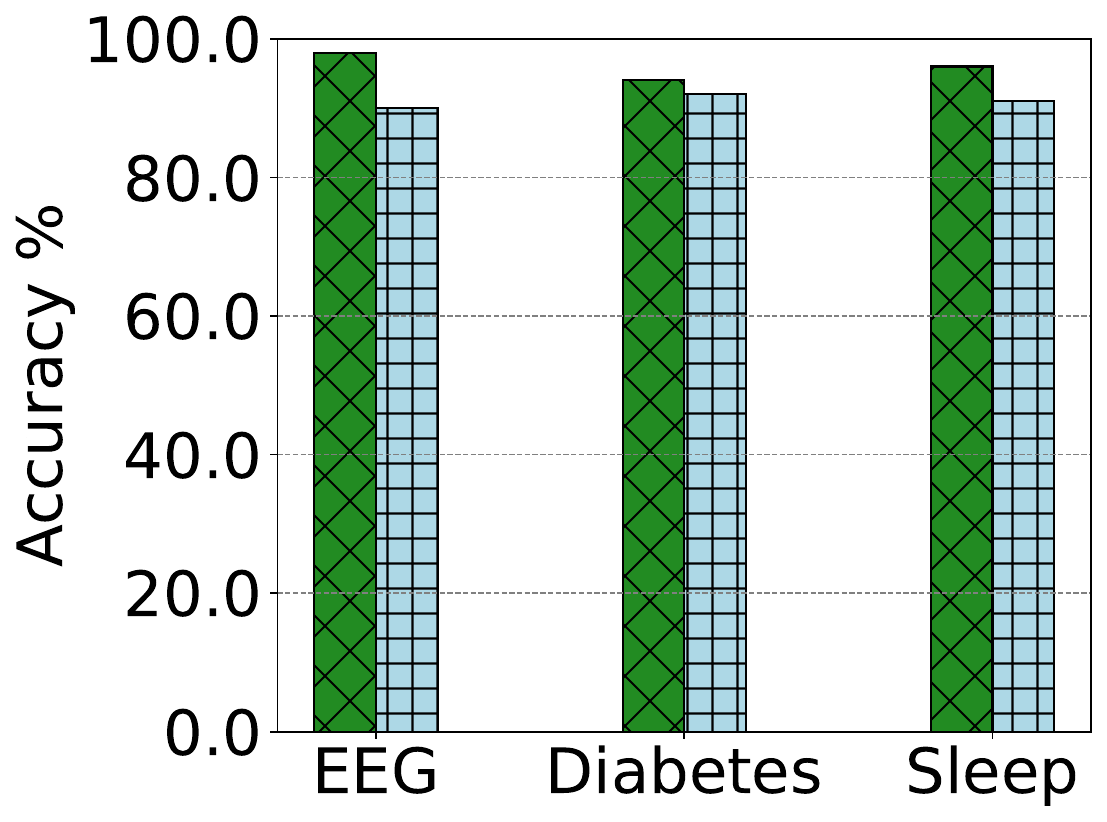}
        \caption{Accuracy with P2W}
        \label{fig:acc_P2W}
    \end{subfigure}
    \hfill
    \begin{subfigure}[b]{0.49\columnwidth}
        \includegraphics[width=\textwidth]{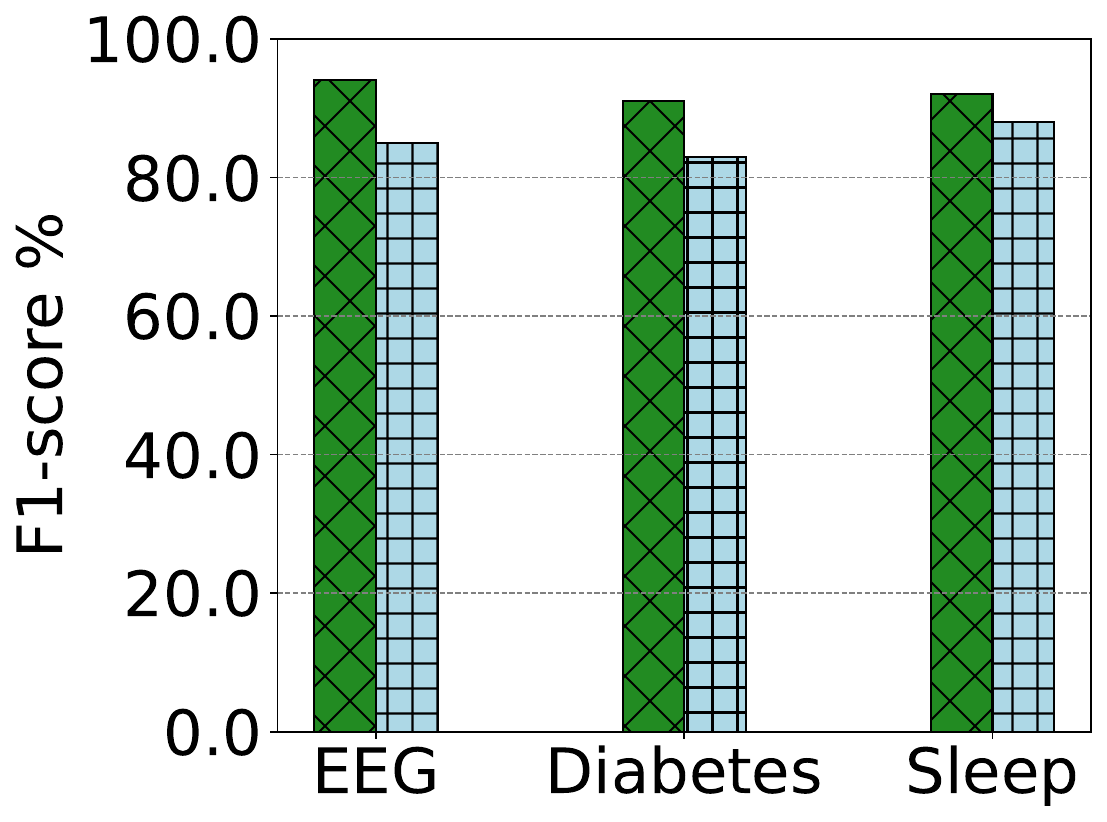}
        \caption{F1-score with P2W}
        \label{fig:f1_score_P2W}
    \end{subfigure}

    \caption{Accuracy and F1-score of ML models obtained with and without P2W using balanced or imbalanced $D_\text{small}$.}
    \label{fig:ACC-F1}
    \vspace{-17pt}
\end{figure}

The imbalanced $D_\text{small}$ datasets for this evaluation are constructed by randomly taking data samples from the large datasets, listed in column \texttt{Ref} of Table~\ref{tab:small_datasets}, such that in the constructed EEG, Diabetes, and Sleep $D_\text{small}$ datasets, the number of samples in one of the class is 2$\times$ greater than in the other classes. We obtain new ML models with P2W that are fine-tuned with the imbalanced $D_\text{small}$ datasets as well as new ML models trained only with the same imbalanced datasets. Then, we check the average accuracy and F1-score of the models. We conduct the same experiments using the balanced $D_\text{small}$ datasets, discussed in Section~\ref{sec:D_small}, and compare all results. Note that we check both the accuracy and F1-score because, unlike the accuracy, which can be misleading if a model performs very well only on one class, the F1-score, as we discussed in Section~\ref{sec:F1}, helps ensure that the model’s predictive performance is reliably evaluated across all classes~\cite{10.1007/11941439_114}. 

The results of the aforementioned experiments are depicted by the four plots in Figure~\ref{fig:ACC-F1}. In all plots, the horizontal axis shows the three classification tasks (EEG, Diabetes, and Sleep) performed by the obtained new ML models, and the vertical axis shows the model's accuracy or F1-score. For each model, there is a pair of bars where the left bar indicates the accuracy or F1-score of the new ML model trained/fine-tuned with balanced $D_\text{small}$ datasets, and the right bar indicates the same for imbalanced $D_\text{small}$ datasets.
Figure~\ref{fig:acc} and \ref{fig:f1_score} show the accuracy and F1-score of the new ML models trained only with the $D_\text{small}$ datasets. Figure~\ref{fig:acc_P2W} and \ref{fig:f1_score_P2W} show the same metrics for the new ML models obtained with P2W including fine-tuning with the same $D_\text{small}$ datasets.

By comparing the white bars with the gray bars (our reference points) in  Figure~\ref{fig:acc} and~\ref{fig:f1_score}, it can be seen that when a new ML model is trained only with a small imbalanced datasets $D_\text{small}$, the model's accuracy and F1-score drop significantly by up to 32\% and 72\%, respectively. However, when our P2W approach is utilized including fine-tuning with the same imbalanced datasets, the accuracy and F1-score drop only by up to 5\% and 7\%, respectively, as indicated in Figure~\ref{fig:acc_P2W} and Figure~\ref{fig:f1_score_P2W} (compare the blue bars with the green bars as reference points).
The results in Figure~\ref{fig:ACC-F1} clearly indicate that the new ML models obtained with P2W not only achieve much higher accuracy and F1-score across all three classification tasks compared to the new ML models obtained without P2W, but they also show that our P2W approach is very robust to imbalanced training data, thereby delivering new ML models with high and unbiased predictive accuracy.

\section{Discussion}
\label{sec:discussion}
In this section, we discuss the limitations and future work concerning our P2W approach.

\textit{Limitations:} 
In the first two paragraphs of Section \ref{sec:P2W}, we explain our assumptions regarding P2W together with real-world examples for each assumption to show that P2W is realistic in several scenarios. However, the P2W approach is not very effective in scenarios where some of these assumptions are not true. For example, in order to deploy P2W, access to the power supply of the target SoC is needed, which might not always be the case. Nevertheless, for embedded devices, it is realistic that potential users are in the vicinity of the device and the power supply or even own the device. Moreover, instead of measuring the power consumption, similar results can be achieved by measuring the electromagnetic emanation of the chip, which can be done at a distance from the chip \cite{10.1007/3-540-48405-1_25}.

Another limitation is that the clone SoC needs to be of the exact same type as the target SoC. For broadly used embedded SoCs, it is not a problem to find suitable clone SoCs, as explained in Section~\ref{sec:P2W}, but for niche applications or exotic devices, it is more difficult to apply P2W.

Finally, P2W requires that there is prior knowledge on the topology of the target ML model. If this is not the case, the technique cannot be applied. To mitigate this limitation, a preliminary power consumption measurement on the target SoC can be performed in order to learn about the type of ML model that is used, as has been shown in related work \cite{10077551, 10.5555/3361338.3361374}. Moreover, for certain applications, it is commonly known which ML topologies are being deployed.

\textit{Future work:} 
We plan to study the effects of the complexity of very large ML models within embedded devices on power traces. Although ML models utilized in embedded SoCs are typically not large, the scalability of the P2W approach to very large models could be a valuable path for future research. Additionally, as we evaluated P2W on SoCs with an ARM processor inside, exploring the effectiveness of P2W on different types of hardware, such as FPGAs and GPUs, is a promising direction for future research. Given the parallel processing nature of FPGAs and GPUs, it is beneficial to study how they affect power traces in reflecting the relationships and patterns related to ML model computations.

\section{Conclusions}
\label{sec:Conclusions}

This paper presents P2W, an unconventional transfer learning approach for ML models. Our approach is useful for scenarios where direct access to the coefficients (weights and biases) of existing ML models is not feasible and there is only a relatively small dataset available to train a new ML model from scratch. P2W translates a power trace, captured from an embedded SoC running a target ML model inside, to a weights matrix by utilizing an encoder-decoder deep neural network. This weights matrix approximates the coefficients of the target ML model and is used to initialize a new ML model, thereby performing transfer learning from the target to the new ML model.

Experimenting with relatively small training datasets, we evaluate the effectiveness of our P2W approach by comparing the accuracy of a new ML model trained
with and without using P2W. 

Our experimental results show that after training the model with our P2W approach, we are able to improve the initial 37\% accuracy of the model, trained with only a relatively small dataset, to 97\% when P2W is utilized including fine-tuning with the same dataset. These results  clearly indicate
that new ML models, obtained by P2W, achieve very high predictive accuracy.

In addition, we perform an evaluation of P2W in extreme scenarios where we only have small imbalanced datasets, i.e., datasets within which one or some of the classes have a much larger number of data samples than the others. The results of this evaluation show that for a new ML model trained with only a small imbalanced dataset, the accuracy and F1-score drop significantly by up to 32\% and 72\%, respectively. However, when P2W is utilized including fine-tuning with the same imbalanced dataset, the accuracy and F1-score drop only by up to 5\% and 7\%, respectively. These results suggest that our P2W approach is very robust to imbalanced training data, thereby delivering new ML models with high and unbiased predictive accuracy.

\bibliography{main} 

\begin{thebibliography}{10}

\bibitem{noauthor_stm32_nodate}
{STM32} {MCU} {Developer} {Zone} - {STMicroelectronics}.

\bibitem{noauthor_stmicroelectronics_2023}
{STMicroelectronics} – {STM32} model zoo, November 2023.
\newblock original-date: 2023-01-10T13:58:28Z.

\bibitem{10.5555/3361338.3361374}
Lejla Batina, Shivam Bhasin, Dirmanto Jap, and Stjepan Picek.
\newblock Csi nn: Reverse engineering of neural network architectures through electromagnetic side channel.
\newblock In {\em Proceedings of the 28th USENIX Conference on Security Symposium}, SEC'19, page 515–532, USA, 2019. USENIX Association.

\bibitem{9399005}
Brian Coffen and Md.Shaad Mahmud.
\newblock Tinydl: Edge computing and deep learning based real-time hand gesture recognition using wearable sensor.
\newblock In {\em 2020 IEEE International Conference on E-health Networking, Application \& Services (HEALTHCOM)}, pages 1--6, 2021.

\bibitem{georgy_blood_nodate}
ZUBKOV GEORGY.
\newblock Blood {Pressure} {Data} for disease {Prediction}.

\bibitem{goodfellow2016deep}
Ian Goodfellow, Yoshua Bengio, and Aaron Courville.
\newblock {\em Deep learning}.
\newblock MIT press, 2016.

\bibitem{10.5555/3104482.3104533}
Maayan Harel and Shie Mannor.
\newblock Learning from multiple outlooks.
\newblock In {\em Proceedings of the 28th International Conference on International Conference on Machine Learning}, ICML'11, page 401–408, Madison, WI, USA, 2011. Omnipress.

\bibitem{Hassan2023}
Esraa Hassan, Mahmoud~Y. Shams, Noha~A. Hikal, and Samir Elmougy.
\newblock The effect of choosing optimizer algorithms to improve computer vision tasks: a comparative study.
\newblock {\em Multimedia Tools and Applications}, 82(11):16591--16633, May 2023.

\bibitem{10.1145/3195970.3196105}
Weizhe Hua, Zhiru Zhang, and G.~Edward Suh.
\newblock Reverse engineering convolutional neural networks through side-channel information leaks.
\newblock In {\em Proceedings of the 55th Annual Design Automation Conference}, DAC '18, New York, NY, USA, 2018. Association for Computing Machinery.

\bibitem{10.1007/978-3-030-01364-6_17}
Amelia Jim{\'e}nez-S{\'a}nchez, Shadi Albarqouni, and Diana Mateus.
\newblock Capsule networks against medical imaging data challenges.
\newblock In {\em Intravascular Imaging and Computer Assisted Stenting and Large-Scale Annotation of Biomedical Data and Expert Label Synthesis}, pages 150--160, Cham, 2018. Springer International Publishing.

\bibitem{Kachuee_2018}
Mohammad Kachuee, Shayan Fazeli, and Majid Sarrafzadeh.
\newblock Ecg heartbeat classification: A deep transferable representation.
\newblock In {\em 2018 IEEE International Conference on Healthcare Informatics (ICHI)}. IEEE, June 2018.

\bibitem{10.1007/3-540-48405-1_25}
Paul Kocher, Joshua Jaffe, and Benjamin Jun.
\newblock Differential power analysis.
\newblock In Michael Wiener, editor, {\em Advances in Cryptology --- CRYPTO' 99}, pages 388--397, Berlin, Heidelberg, 1999. Springer Berlin Heidelberg.

\bibitem{10.1109/CVPR.2011.5995702}
B.~Kulis, K.~Saenko, and T.~Darrell.
\newblock What you saw is not what you get: Domain adaptation using asymmetric kernel transforms.
\newblock In {\em Proceedings of the 2011 IEEE Conference on Computer Vision and Pattern Recognition}, CVPR '11, page 1785–1792, USA, 2011. IEEE Computer Society.

\bibitem{laksika_tharmalingam_sleep_nodate}
{Laksika Tharmalingam}.
\newblock Sleep {Health} and {Lifestyle} {Dataset}.

\bibitem{LEE202015}
Shin-Jye Lee, Ching-Hsun Tseng, G.T.–R. Lin, Yun Yang, Po~Yang, Khan Muhammad, and Hari~Mohan Pandey.
\newblock A dimension-reduction based multilayer perception method for supporting the medical decision making.
\newblock {\em Pattern Recognition Letters}, 131:15--22, 2020.

\bibitem{LI2023107290}
Penghui Li, Rui Zhou, Jin He, Shifeng Zhao, and Yun Tian.
\newblock A global-frequency-domain network for medical image segmentation.
\newblock {\em Computers in Biology and Medicine}, 164:107290, 2023.

\bibitem{electronics11162545}
Ioan Lucan~Orășan, Ciprian Seiculescu, and Cătălin~Daniel Căleanu.
\newblock A brief review of deep neural network implementations for arm cortex-m processor.
\newblock {\em Electronics}, 11(16), 2022.

\bibitem{noauthor_evaluation_nodate}
{microchip}.
\newblock Evaluation {Boards} {\textbar} {Microchip} {Technology}.

\bibitem{10.1145/2786805.2786814}
Jaechang Nam and Sunghun Kim.
\newblock Heterogeneous defect prediction.
\newblock In {\em Proceedings of the 2015 10th Joint Meeting on Foundations of Software Engineering}, ESEC/FSE 2015, page 508–519, New York, NY, USA, 2015. Association for Computing Machinery.

\bibitem{nielsen2015neural}
Michael~A Nielsen.
\newblock {\em Neural networks and deep learning}, volume~25.
\newblock Determination press San Francisco, CA, USA, 2015.

\bibitem{o2014chipwhisperer}
Colin O’flynn and Zhizhang Chen.
\newblock Chipwhisperer: An open-source platform for hardware embedded security research.
\newblock In {\em Constructive Side-Channel Analysis and Secure Design: 5th International Workshop, COSADE 2014, Paris, France, April 13-15, 2014. Revised Selected Papers 5}, pages 243--260. Springer, 2014.

\bibitem{5288526}
Sinno~Jialin Pan and Qiang Yang.
\newblock A survey on transfer learning.
\newblock {\em IEEE Transactions on Knowledge and Data Engineering}, 22(10):1345--1359, 2010.

\bibitem{10.5555/3454287.3455008}
Adam Paszke, Sam Gross, Francisco Massa, Adam Lerer, James Bradbury, Gregory Chanan, Trevor Killeen, Zeming Lin, Natalia Gimelshein, Luca Antiga, Alban Desmaison, Andreas K\"{o}pf, Edward Yang, Zach DeVito, Martin Raison, Alykhan Tejani, Sasank Chilamkurthy, Benoit Steiner, Lu~Fang, Junjie Bai, and Soumith Chintala.
\newblock {\em PyTorch: An Imperative Style, High-Performance Deep Learning Library}.
\newblock Curran Associates Inc., Red Hook, NY, USA, 2019.

\bibitem{pavan_bodanki_blood_nodate}
{PAVAN BODANKI}.
\newblock Blood {Pressure} {Data} for disease {Prediction}.

\bibitem{noauthor_plumerai_nodate}
{Plumerai}.
\newblock Plumerai wins {MLPerf} {Tiny} 1.1 {AI} benchmark for microcontrollers again {\textbar} {Plumerai} {Blog}.

\bibitem{Rana_2023}
Tanzeel~Sultan Rana, Imran Saleem, Rabia~Naseer Rao, Maryam Shabbir, and Laiba~Wahid Chaudhry.
\newblock Comparative analysis of breast cancer detection using cutting-edge machine learning algorithms (mlas).
\newblock {\em Innovative Computing Review}, 3(1), Jun. 2023.

\bibitem{Real2021PhysicalSA}
Maria~M{\'e}ndez Real and Rub{\'e}n Salvador.
\newblock Physical side-channel attacks on embedded neural networks: A survey.
\newblock {\em ArXiv}, abs/2110.11290, 2021.

\bibitem{Sarker2023}
Iqbal~H. Sarker, Asif~Irshad Khan, Yoosef~B. Abushark, and Fawaz Alsolami.
\newblock Internet of things (iot) security intelligence: A comprehensive overview, machine learning solutions and research directions.
\newblock {\em Mobile Networks and Applications}, 28(1):296--312, Feb 2023.

\bibitem{9669005}
Gawsalyan Sivapalan, Koushik~Kumar Nundy, Soumyabrata Dev, Barry Cardiff, and Deepu John.
\newblock Annet: A lightweight neural network for ecg anomaly detection in iot edge sensors.
\newblock {\em IEEE Transactions on Biomedical Circuits and Systems}, 16(1):24--35, 2022.

\bibitem{Smith1988-ux}
J~W Smith, J~E Everhart, W~C Dickson, W~C Knowler, and R~S Johannes.
\newblock Using the {ADAP} learning algorithm to forecast the onset of diabetes mellitus.
\newblock In {\em Proceedings of the Annual Symposium on Computer Application in Medical Care}, pages 261--265. 1988.

\bibitem{10.1007/11941439_114}
Marina Sokolova, Nathalie Japkowicz, and Stan Szpakowicz.
\newblock Beyond accuracy, f-score and roc: A family of discriminant measures for performance evaluation.
\newblock In Abdul Sattar and Byeong-ho Kang, editors, {\em AI 2006: Advances in Artificial Intelligence}, pages 1015--1021, Berlin, Heidelberg, 2006. Springer Berlin Heidelberg.

\bibitem{Taha2015}
Abdel~Aziz Taha and Allan Hanbury.
\newblock Metrics for evaluating 3d medical image segmentation: analysis, selection, and tool.
\newblock {\em BMC Medical Imaging}, 15(1):29, Aug 2015.

\bibitem{tharwat2016principal}
Alaa Tharwat.
\newblock Principal component analysis-a tutorial.
\newblock {\em International Journal of Applied Pattern Recognition}, 3(3):197--240, 2016.

\bibitem{10077551}
Ziyu Wang, Fan-hsuan Meng, Yongmo Park, Jason~K. Eshraghian, and Wei~D. Lu.
\newblock Side-channel attack analysis on in-memory computing architectures.
\newblock {\em IEEE Transactions on Emerging Topics in Computing}, pages 1--13, 2023.

\bibitem{warden2019tinyml}
Pete Warden and Daniel Situnayake.
\newblock {\em Tinyml: Machine learning with tensorflow lite on arduino and ultra-low-power microcontrollers}.
\newblock O'Reilly Media, 2019.

\bibitem{Weiss2016}
Karl Weiss, Taghi~M. Khoshgoftaar, and DingDing Wang.
\newblock A survey of transfer learning.
\newblock {\em Journal of Big Data}, 3(1):9, May 2016.

\bibitem{9000972}
Yun Xiang, Zhuangzhi Chen, Zuohui Chen, Zebin Fang, Haiyang Hao, Jinyin Chen, Yi~Liu, Zhefu Wu, Qi~Xuan, and Xiaoniu Yang.
\newblock Open dnn box by power side-channel attack.
\newblock {\em IEEE Transactions on Circuits and Systems II: Express Briefs}, 67(11):2717--2721, 2020.

\bibitem{8735505}
Kota Yoshida, Takaya Kubota, Mitsuru Shiozaki, and Takeshi Fujino.
\newblock Model-extraction attack against fpga-dnn accelerator utilizing correlation electromagnetic analysis.
\newblock In {\em 2019 IEEE 27th Annual International Symposium on Field-Programmable Custom Computing Machines (FCCM)}, pages 318--318, 2019.

\bibitem{Zhou_Pan_Tsang_Yan_2014}
Joey Zhou, Sinno Pan, Ivor Tsang, and Yan Yan.
\newblock Hybrid heterogeneous transfer learning through deep learning.
\newblock {\em Proceedings of the AAAI Conference on Artificial Intelligence}, 28(1), Jun. 2014.

\bibitem{JMLR:v20:13-580}
Joey~Tianyi Zhou, Ivor~W. Tsang, Sinno~Jialin Pan, and Mingkui Tan.
\newblock Multi-class heterogeneous domain adaptation.
\newblock {\em Journal of Machine Learning Research}, 20(57):1--31, 2019.

\bibitem{Zhu_Chen_Lu_Pan_Xue_Yu_Yang_2011}
Yin Zhu, Yuqiang Chen, Zhongqi Lu, Sinno Pan, Gui-Rong Xue, Yong Yu, and Qiang Yang.
\newblock Heterogeneous transfer learning for image classification.
\newblock {\em Proceedings of the AAAI Conference on Artificial Intelligence}, 25(1):1304--1309, Aug. 2011.

\bibitem{10172347}
Zhuangdi Zhu, Kaixiang Lin, Anil~K. Jain, and Jiayu Zhou.
\newblock Transfer learning in deep reinforcement learning: A survey.
\newblock {\em IEEE Transactions on Pattern Analysis and Machine Intelligence}, 45(11):13344--13362, 2023.

\bibitem{data4010014}
Igor Zyma, Sergii Tukaev, Ivan Seleznov, Ken Kiyono, Anton Popov, Mariia Chernykh, and Oleksii Shpenkov.
\newblock Electroencephalograms during mental arithmetic task performance.
\newblock {\em Data}, 4(1), 2019.

\end{thebibliography}
\bibliographystyle{plain}  

\end{document}